\documentclass[10pt,twocolumn,letterpaper]{article}
\usepackage{cvpr}
\usepackage[dvipsnames]{xcolor}
\usepackage{times}
\usepackage{amsmath}
\usepackage{amssymb}
\usepackage{arydshln}
\usepackage{bm}
\usepackage{boldline}
\usepackage{booktabs}
\usepackage{color}
\usepackage{enumitem}
\usepackage{float}
\usepackage{graphicx}
\usepackage{multirow}
\usepackage{nicefrac}
\usepackage[accsupp]{axessibility}

\definecolor{cvprblue}{rgb}{0.21,0.49,0.74}
\usepackage[pagebackref=true,breaklinks=true,colorlinks,citecolor=cvprblue]{hyperref}
\usepackage[capitalize]{cleveref}

\newcommand{\method}{Free3D\xspace}

\makeatletter
\renewcommand{\paragraph}{%
    \@startsection{paragraph}{4}%
    {\z@}{-0.5em}{-0.5em}%
    {\normalfont\normalsize\bfseries}%
}
\makeatother

\def\paperID{737} 
\def\confName{CVPR}
\def\confYear{2024}

\title{\method: Consistent Novel View Synthesis without 3D Representation}
\author{
Chuanxia Zheng \quad Andrea Vedaldi \\[0.3em]
Visual Geometry Group, University of Oxford \\
{\tt\small \{cxzheng, vedaldi\}@robots.ox.ac.uk}}

\begin{document}
\twocolumn[
{
\maketitle
\vspace{-1cm}
\begin{figure}[H]
    \centering
    \hsize=\textwidth
    \includegraphics[width=\textwidth]{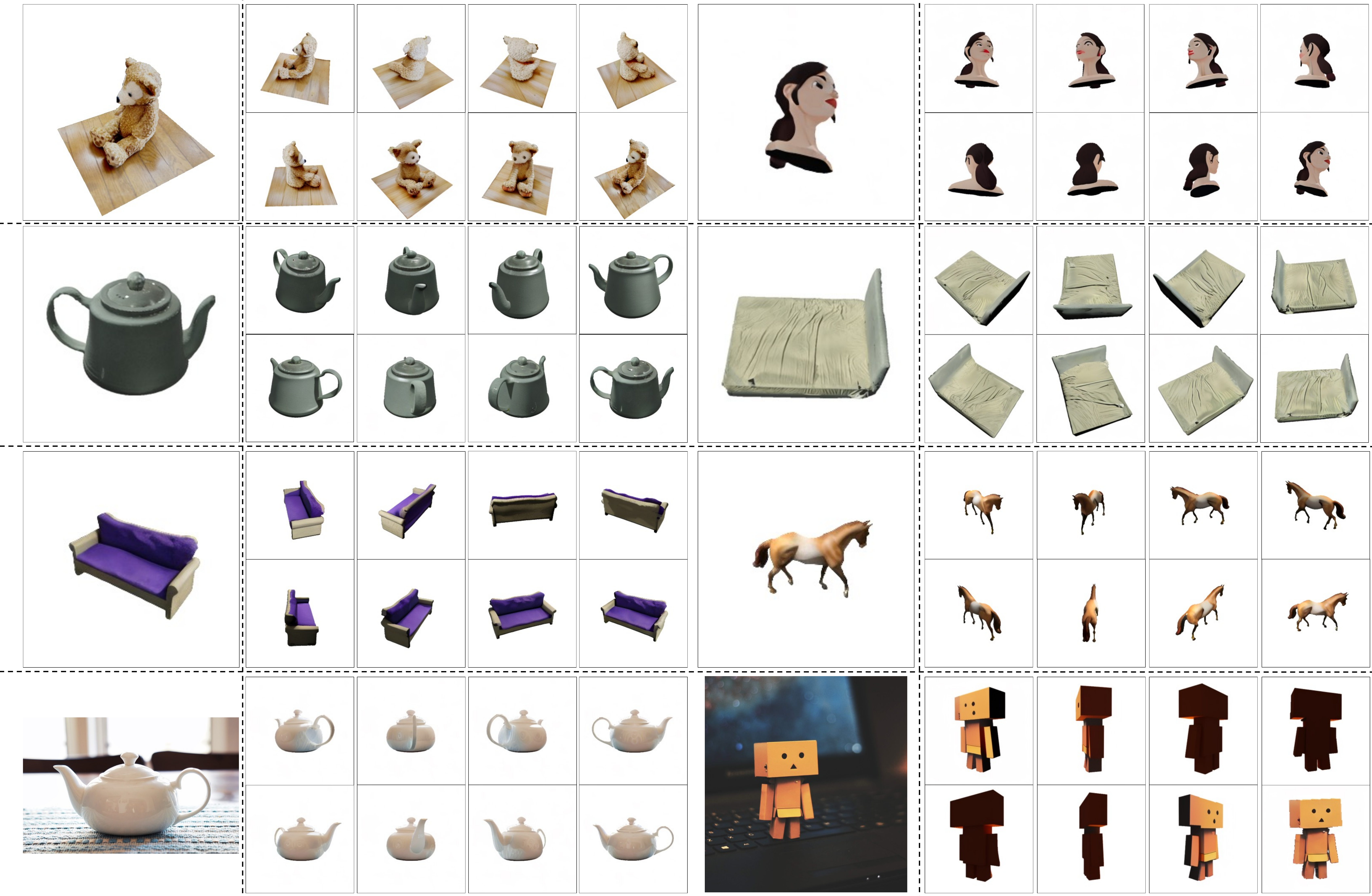}
    \begin{picture}(0,0)
    \put(-248,270){\rotatebox{90}{\footnotesize Objaverse~\cite{deitke2023objaverse}}}
    \put(-248,182){\rotatebox{90}{\footnotesize OmniObject3D~\cite{wu2023omniobject3d}}}
    \put(-248,118){\rotatebox{90}{\footnotesize GSO~\cite{downs2022google}}}
    \put(-248,32){\rotatebox{90}{\footnotesize Real Images}}
    \put(-218, 4){\footnotesize Input View}
    \put(-132,4){\footnotesize Generated Views by our \method}
    \put(28, 4){\footnotesize Input View}
    \put(117,4){\footnotesize Generated Views by our \method}
    \end{picture}
    \vspace{-0.2cm}
    \caption{
    Given a single input view, \textbf{\method} synthesizes consistent $360^\circ$ views accurately without using an explicit 3D representation.
    Trained on \texttt{Objaverse} only, it generalizes well to new datasets and categories.}%
    \label{fig:teaser}
\end{figure}
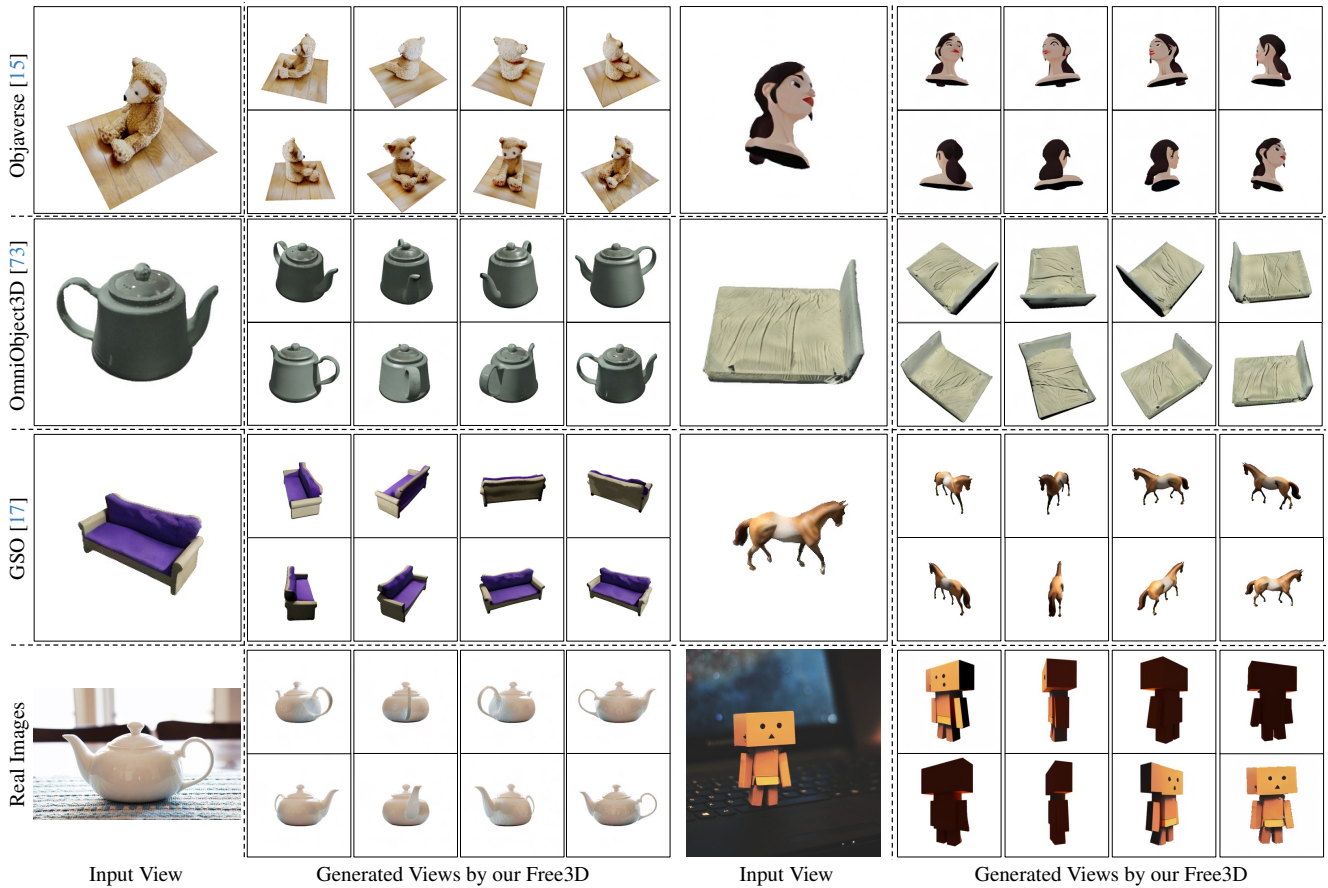
}
]

\maketitle
\begin{abstract}
\vspace{-0.6em}
We introduce \method, a simple accurate method for monocular open-set novel view synthesis (NVS).
Similar to Zero-1-to-3, we start from a pre-trained 2D image generator for generalization, and fine-tune it for NVS\@.
Compared to other works that took a similar approach, we obtain significant improvements without resorting to an explicit 3D representation, which is slow and memory-consuming, and without training an additional network for 3D reconstruction.
Our key contribution is to improve the way the target camera pose is encoded in the network, which we do by introducing a new ray conditioning normalization (RCN) layer.
The latter injects pose information in the underlying 2D image generator by telling each pixel its viewing direction.
We further improve multi-view consistency by using light-weight multi-view attention layers and by sharing generation noise between the different views.
We train \method on the Objaverse dataset and demonstrate excellent generalization to new categories in new datasets, including OmniObject3D and GSO\@.
The project page is available at \href{https://chuanxiaz.com/free3d/}{https://chuanxiaz.com/free3d/}.
\end{abstract}
\section{Introduction}%
\label{sec:intro}

Novel view synthesis (NVS) has seen significant recent progress~\cite{barron2021mip,muller2022instant,barron2022mip,chen2022tensorf,kerbl20233d}, in part due to new neural representations like NeRF~\cite{mildenhall2020nerf}.
However, many such NVS methods require to optimize a 3D model from scratch \emph{for each scene}, and require dozens of input views to work well; they are thus impractical in many applications.
This has motivated authors to apply generative models~\cite{kingma2013auto,goodfellow2014generative,sohl2015deep} to the NVS task, eschewing the need for an explicit 3D model and enabling NVS from a single image~\cite{park2017transformation,kanazawa2018learning,yu2021pixelnerf,huang2022planes,watson2022novel,melas2023pc2,szymanowicz23viewset}.
Most of these learn 3D priors that are applicable to an entire \emph{object category}, or even to \emph{unstructured collections of objects}~\cite{park2017transformation,kanazawa2018learning,yu2021pixelnerf,huang2022planes,watson2022novel,melas2023pc2,szymanowicz23viewset}.
In this work, we aim at improving the quality of these approaches while also generalizing further their applicability by considering an \emph{open-set} setting, where at test time one is given not only new object instances and categories, but also new datasets.

There are two quality targets for NVS\@:
\textbf{(i)} The output must accurately reflect the pose of the target cameras,
and \textbf{(ii)}, when several views of the same object are generated, they must be mutually consistent.
In order to achieve these goals, recent methods~\cite{liu2023one,liu2023syncdreamer,shi2023mvdream,yang2023consistnet,xu2024grm} build a 3D representation of the object or scene, often combined with a pre-trained 2D generative model~\cite{rombach2022high}.
Using a 2D model with a 3D representation works well but adds complexity.

In this paper, we introduce \method, a simple and efficient method that can also achieve consistent NVS results
\emph{without the need to rely on an explicit 3D representation}.
Zero-1-to-3~\cite{liu2023zero} and its precursor 3DiM~\cite{watson2022novel} are perhaps the best-known examples of such a 3D-free NVS system.
Like Zero-1-to-3, \method builds upon a pre-trained 2D generative model like Stable Diffusion~\cite{rombach2022high}, trained on a large-scale dataset (\ie LAION 5B~\cite{schuhmann2022laion}), as a data prior.
The prior knowledge contained in such a 2D generator is extremely important to be able to `guess' plausible novel views of open-set objects, which is inherently highly ambiguous.
However, we show empirically that Zero-1-to-3 has, in practice, poor camera pose control,
and, when tasked with generating multiple views, not very consistent.
The latter is unavoidable in their design because each view is sampled independently from scratch and, owing to the ambiguity of reconstruction, there is no reason why compatible images would be generated each time.

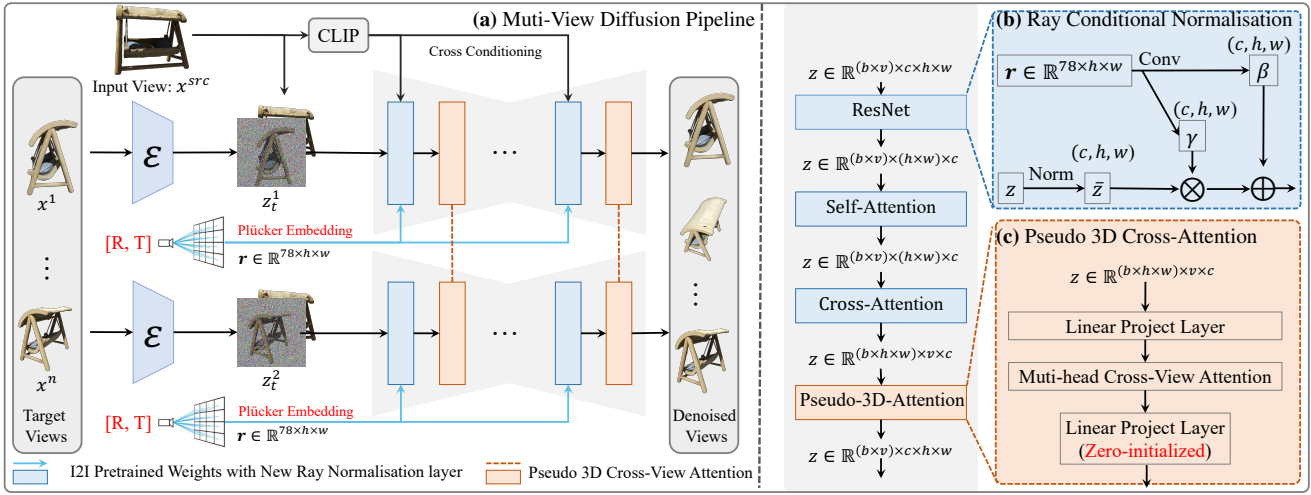
\begin{figure*}[tb!]
\centering
\includegraphics[width=\linewidth]{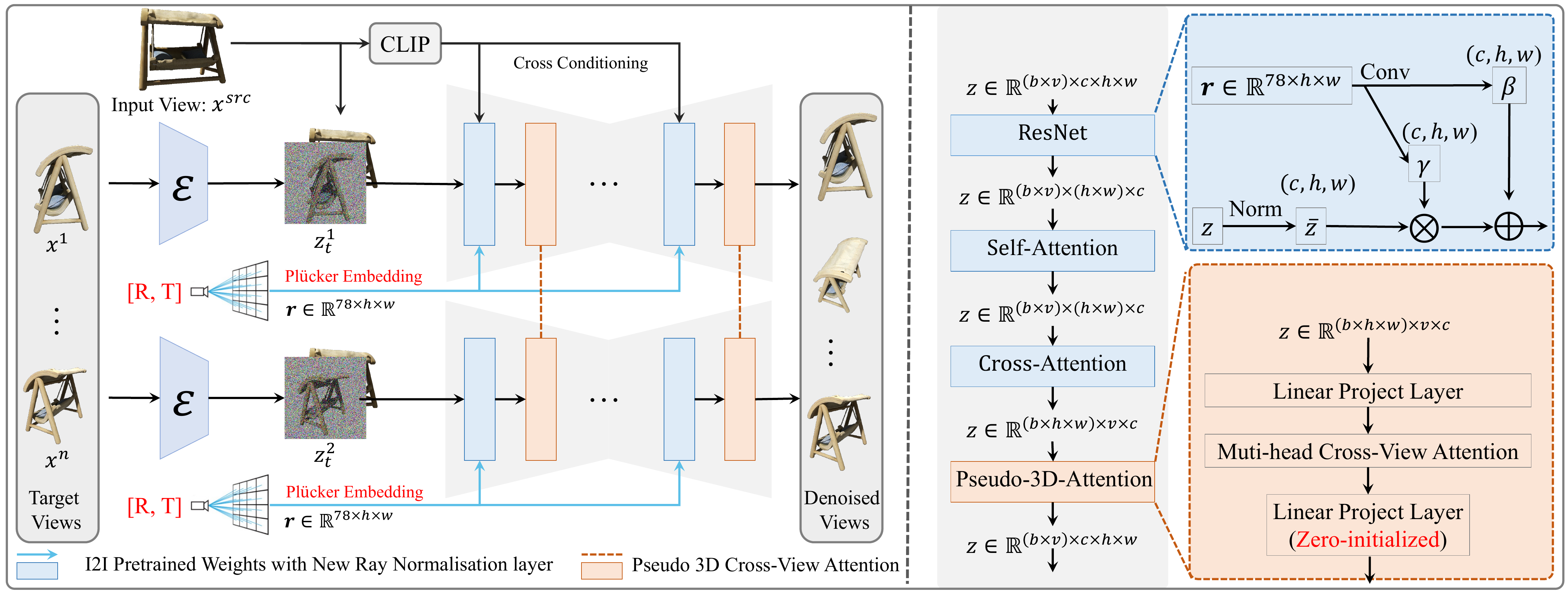}
\begin{picture}(0,0)
\put(-68,190){\footnotesize \textbf{(a)} Muti-View Diffusion Pipeline}
\put(130,190){\footnotesize \textbf{(b)} Ray Conditional Normalisation}
\put(130,108){\footnotesize \textbf{(c)} Pseudo 3D Cross-Attention}
\end{picture}
\vspace{-15pt}
\caption{\textbf{The overall pipeline of our \method.}
(a) Given a single input image, the \method jointly predicts multiple target views, instead of processing them independently.
(b) We propose a novel \emph{ray conditional normalization} (RCN) layer, which uses a \emph{per-pixel} oriented camera ray to module the latent features, enabling the model's ability to capture more precise viewpoints.
(c) A memory-friendly \emph{pseudo-3D cross-attention} module is introduced to efficiently bridge information across multiple generated views. 
}\label{fig:framework}
\end{figure*}

To mitigate these issues, we first show that better camera control can be achieved by switching to a different representation of camera pose.
Specifically, we introduce a \emph{ray conditioning normalization} (RCN) layer (\cref{fig:framework} (b)) which tells each pixel its viewing direction.
This is a distributed representation of the camera, which should be contrasted to the concentrated camera representation used in~\cite{liu2023zero}, where the camera pose is processed as language-like tokens that may be difficult for the network to interpret and utilize~\cite{sargent2023zeronvs}.
In contrast, with our RCN layer, we show how to effectively incorporate this per-pixel ray information in an \emph{existing} text-to-image diffusion model, which empirically leads to significantly more accurate NVS in our experiments (\cref{tab:main_sota,tab:general_sota}).
RCN is inspired by the design of methods like NeRF~\cite{mildenhall2020nerf}, LFNs~\cite{sitzmann2021light} and 3DiM~\cite{watson2022novel}, which also work by modelling individual rays.

While RCN allows to control the camera more accurately, it does not improve multi-view consistency.
For the latter, we introduce a \emph{pseudo-3D cross-view attention} module (\cref{fig:framework}~(c)) which is inspired by video diffusion models~\cite{singer2022make,ho2022imagen,blattmann2023align,guo2023animatediff,wu2023lamp,blattmann2023stable}.
This layer fuses information across all views instead of processing them independently.
Furthermore, we use \emph{multi-view noise sharing} when generating the different views of the object, which further enhances consistency due to the continuity of the denoising function, reducing aleatoric variations between the views.

We benchmark \method against state-of-the-art methods~\cite{liu2023zero,liu2023syncdreamer,weng2023consistent123,objaverseXL} on \emph{open-set} NVS\@.
Although the model is trained on only one dataset, it generalizes well to all recent NVS benchmark datasets, including \texttt{Objaverse}~\cite{deitke2023objaverse}, \texttt{OmniObject3D}~\cite{wu2023omniobject3d}, and Google Scanned Object (\texttt{GSO})~\cite{downs2022google}.
A thorough experimental assessment shows that our approach consistently outperforms existing methods, both quantitatively and qualitatively.

To summarise, with \method we make the following contributions:
(i) We introduce the \emph{ray conditioning normalization} (RCN) layer and show that representing the camera by utilizing a combination of distributed ray conditioning and concentrated pose tokens significantly improves pose accuracy in NVS\@.
(ii) We show that a small \emph{multi-view attention} module is sufficient to improve multi-view consistency by exchanging information between views at a low cost.
(iii) We find that \emph{multi-view noise sharing} further improves consistency.
(iv) We demonstrate empirically that \method achieves consistent NVS \emph{without needing a 3D representation} and outperforms the existing state-of-the-art models on both pose accuracy and view consistency.
\section{Related Work}%
\label{sec:rel}

\paragraph{{Per-Scene} NVS.}

Early NVS works relied on epipolar geometry to interpolate between different views of the same scene~\cite{chen1993view,debec1996modeling}.
A recent breakthrough was to represent 3D scenes as implicit neural fields, as proposed by SRN~\cite{sitzmann2019scene}, DeepSDF~\cite{park2019deepsdf}, NeRF~\cite{mildenhall2020nerf} and LFN~\cite{sitzmann2021light}, and further improved in follow-ups~\cite{zhang2020nerf++,barron2021mip,barron2022mip,fridovich2022plenoxels,chen2022tensorf,muller2022instant,kerbl20233d}.
Even so, data efficiency, generalizability, and robustness remain a limitation: such systems require multiple views to learn the 3D representation from scratch for every scene.

To bypass the need for multiple input views, DreamFusion~\cite{poole2023dreamfusion} proposes to distill 3D models from a large-scale pre-trained 2D diffusion model~\cite{saharia2022photorealistic}.
RealFusion~\cite{melas2023realfusion} extends the latter to single-view image reconstruction by adding the input image as a constraint during distillation.
Several follow-up works~\cite{tang2023make,qian2023magic123,sargent2023zeronvs,purushwalkam2023conrad} further improve the resolution and quality of the resulting 3D assets.
Although these models are open-ended~\cite{melas2023realfusion}, they still require lengthy per-scene optimisation.

\paragraph{{Category-Specific} NVS.}

Some recent work~\cite{tatarchenko2016multi,park2017transformation,wiles2020synsin,chen2023explicit} built the autoencoder architecture for NVS\@.
Driven by the effectiveness of light field rendering~\cite{adelson1991plenoptic,adelson1992single}, other NVS approaches~\cite{suhail2022light,sajjadi2022scene,sajjadi2022object} query a network for colour of different rays.
3DiM~\cite{watson2022novel} then introduced diffusion models into NVS\@, an approach also taken by
RenderDiffusion~\cite{anciukevivcius2023renderdiffusion},
HoloDiffusion~\cite{karnewar2023holodiffusion},
SparseFusion~\cite{zhou2023sparsefusion},
GeNVS~\cite{chan2023genvs},
Viewset Diffusion~\cite{szymanowicz23viewset} and
LFD~\cite{xiong2023light}.
However, these methods train their models from scratch using pure 3D data, which are too small to afford open-set generalization, and are thus limited to one or few categories.

\paragraph{{Open-set} NVS.}

To operate in an open-set setting, Zero-1-to-3~\cite{liu2023zero} builds on a large pre-trained 2D image generator and fine-tune it on the Objaverse dataset~\cite{deitke2023objaverse}.
While it has good generalizability, it fails to achieve high pose accuracy, and its reconstructions are inconsistent across views.
To mitigate these issues, other concurrent approaches either integrate 3D representations into the network~\cite{yang2023consistnet,liu2023syncdreamer,kant2023invs} or train another 3D network~\cite{liu2023one,weng2023consistent123,shi2023mvdream,liu2023one2345++,xu2024grm}.
These methods can output high-quality target views, but are computationally expensive.
More importantly, these methods do \emph{not} address the issue of pose representation, which we find to be a key bottleneck in NVS\@.
In contrast, our method is \emph{3D-free} and achieves comparable or better NVS quality, due to \emph{ray conditioning normalisation}, \emph{multi-view attention}, and \emph{multi-view noise sharing}.
A more recent concurrent work is SVD~\cite{blattmann2023stable}, which synthesizes multi-view videos, but is trained with fixed camera poses.
In contrast, \method is trained to generate arbitrary 360$^\circ$ viewpoints.
The follow-up work SV3D~\cite{voleti2024sv3d} builds on a similar concept.

\section{Method}%
\label{sec:methods}

\newcommand{\bP}{\mathbf{P}}
\newcommand{\bR}{\mathbf{R}}
\newcommand{\bK}{\mathbf{K}}
\newcommand{\bT}{\mathbf{T}}
\newcommand{\bo}{\boldsymbol{o}}
\newcommand{\bd}{\boldsymbol{d}}
\newcommand{\br}{\boldsymbol{r}}

Our goal is to learn a model $\Phi$ that, given as input an image $x^\text{src}$ and a sequence of camera poses
$
\mathcal{P}
=
\{
\bP^i
\}^{N}_{i=1}
$,
synthesizes corresponding novel views
$
\{x^i\}_{i=1}^N
$
which are accurate and consistent without relying on an explicit 3D representation.
We generate all the views together, conditioned on the given inputs, so that the network has a chance to produce several consistent views together.

Specifically, we address two important challenges:
\textbf{(i)} ensuring that the model accurately captures the pose of the target view and
\textbf{(ii)} ensuring the different views consistent in terms of geometry and appearance.
To achieve these, our framework, illustrated in~\cref{fig:framework}, extends a 2D generator by injecting at each layer a ray-conditional normalization layer (\cref{s:pose}) and a pseudo-3D attention layer (\cref{s:autoregressive}).
It also utilizes multi-view noise sharing (\cref{s:autoregressive}).
The former captures pose more accurately, whereas the last two improve multi-view consistency.
{
Although 3DiM~\cite{watson2022novel} also passes ray information to the network, the key difference is that we show how to inject ray conditioning in a \emph{pre-trained} 2D generator.
The benefit is to marry generalization with high pose accuracy without an explicit 3D representation or an additional 3D reconstruction network.
}

\subsection{Ray Conditioning Normalization (RCN)}%
\label{s:pose}

\paragraph{Ray Conditioning Embedding.}

Given a target view
$
\bP = (\bK, \bR, \bT)
$,
for each pixel $(u,v)$ in the image, we define the Pl{\"u}cker coordinates
$
\br_{uv} =
\phi(\bo,\bd_{uv}) =
(\bo \times \bd_{uv}, \bd_{uv})
\in \mathbb{R}^6
$
of the ray going from the camera center
$
\bo \in \mathbb{R}^3
$
through the pixel, where
$
\bd_{uv}
=
\bR^{\top}(\bK^{-1}(u, v, 1)^{\top}-\bT)
\in \mathbb{R}^3
$
is the ray direction and
$
(\bK, \bR, \bT)
$
are the camera's calibration, rotation and translation parameters.
This encoding was originally introduced by LFN~\cite{sitzmann2021light}.
It is invariant to shifting the camera along the ray, meaning that
$
\phi(\bo + \lambda \bd,\bd) =
((\bo + \lambda \bd) \times \bd_{uv}, \bd_{uv}) =
(\bo \times \bd_{uv}, \bd_{uv}) =
\phi(\bo,\bd)
$,
which matches the fact that light propagates in straight lines.

\paragraph{Ray Conditioning Architectures.}

Our goal is to modify the denoising neural network
$
\hat \epsilon(z_,t,y)
$
so that the conditioning information $y$ includes ray conditioning.
We experiment with a number of different architectures to do so, proposing various ray conditioning layers:
\begin{itemize}[itemsep=0pt]
\item \emph{Concatenation.}
Following Zero-1-to-3~\cite{liu2023zero}, a natural choice is to concatenate the noised target $z_t^\text{tgt}$, original source embedding $z^\text{src}$, and ray conditioning at the input, which is then $(z_t^\text{tgt},z^\text{src},\br)$ instead of $z_t^\text{tgt}$ alone.

\item \emph{Multi-scales concatenation.}
We further consider concatenating the ray embeddings $\br$ to each intermediate layer in the UNet $\epsilon_\theta$.
Note that each layer operates at a different resolution, so this amounts to injecting the information $\br$ at different scales.

\item \emph{Ray conditioning normalization (RCN).}
Finally, we propose to combine the adaptive layer norm~\cite{dumoulin2016learned,huang2017arbitrary,karras2019style,zheng2022movq} with ray conditioning to modulate the image latents.
Specifically, the activation latent $F_i$ of the $i$-th layer in the UNet $\epsilon_\theta$ is modulated by:
\begin{equation}
\operatorname{ModLN}_{\br}(F_i)
=
\operatorname{LN}(F_i) \cdot (1 + \gamma) + \beta,
\end{equation}
where
$
(\gamma, \beta) = \operatorname{MLP}^\text{mod}(\br)
$
are \emph{scale} and \emph{shift} parameters predicted from the ray embeddings $\br$.
This is applied to each sub-module of the UNet $\epsilon_\theta$ (\cref{fig:framework} (a) \& (b)).
\end{itemize}
Interestingly, while RCN works best, we show empirically that all such architectures lead to strong improvements (\cref{tab:config_abltion}), which confirms that the key factor in better camera control is to inject ray information in the network.

\paragraph{Discussion.}

RCN differs significantly from Zero-1-to-3~\cite{liu2023zero} and follow-ups~\cite{liu2023one,liu2023syncdreamer,yang2023consistnet,weng2023consistent123,jiang2023efficient,voleti2024sv3d} which encode the camera pose $\bP$ as \emph{global} tokens.
Ray-Cond~\cite{chen2023:ray-conditioning} and LFD~\cite{xiong2023light} also applied ray conditioning for NVS in GAN and Diffusion, respectively, but they only concatenate it as additional channels, and consider category-specific NVS\@.
Similar to  \emph{ray-casting}~\cite{lombardi2019neural,mildenhall2020nerf}, RCN considers one ray per pixel, but, differently from the, does \emph{not} evaluate hundreds of samples per ray.
Because of this, \method dramatically reduces the rendering time and memory consumption compared to concurrent works like~\cite{liu2023syncdreamer,liu2023one,yang2023consistnet}.

\subsection{View Consistent Rendering}%
\label{s:autoregressive}

Given a source image $x^\text{src}$, our goal is to render a series of consistent novel views $x^i$.
While the camera encoding technique of~\cref{s:pose} significantly enhances pose accuracy, if images are sampled independently, they will almost never be visually consistent due to the intrinsic ambiguity of the reconstruction task.
To remove or at least greatly mitigate this problem, we propose to sample images jointly.

\paragraph{Multi-view attention.}

We first adapt the frame attention, a well-established method in video generators~\cite{singer2022make,ho2022imagen,blattmann2023align,guo2023animatediff,wu2023lamp,blattmann2023stable}, to capture temporal dependencies across
views.
As shown in \cref{fig:framework}~(c), given a 5D latent
$
z\in\mathbb{R}^{B\times v \times c\times h\times w}
$,
we initially reshape it to
$
z\in\mathbb{R}^{(B\times h \times w) \times v \times c}
$,
resulting in $batch\times height\times width$ sequences at the length of $views$.
Subsequently, this reshaped latent is passed through the pseudo-3D attention module to calculate the similarity across different views.
Since this attention layer operates across views but separately for each spatial location, the computational and memory costs are quite low (\cref{tab:main_sota}).
Similar to ray conditioning, we inject multi-view attention at each level of the UNet $\hat\epsilon_\theta$.

\paragraph{Multi-view noise sharing.}

In order to reduce the variance between different views, we propose to start sampling each view from the \emph{same} noise vector $x_T$.
The network $\epsilon_\theta$ still generates different views because it is conditioned on the camera parameters.
It can also generate different reconstructions by taking a new noise sample~\cref{fig:diverse}.
However, sharing noise reduces the variance \emph{between views}.
This can be justified by noting that the network $\hat\epsilon_\theta(z_t,t,y)$ is a continuous function of both $z_t$ and $y$~\cite{wu2023uncovering}.
Besides, different from~\cite{tseng2023consistent},
we do not specially set the noise schedule in different time steps $t$.
More complex designs and training strategies have the potential to improve performance but are not the focus of this work.

\subsection{Learning formulation}

The learning objective is given by
\begin{equation}\label{eq:loss_lg}
    \mathcal{L} =
    \mathbb{E}_{(Z_0,z^\text{src},\bP),\epsilon,t}
    \left[\lVert
        \epsilon - \epsilon_\theta(Z_t, t, y)
    \rVert^2_2\right],
\end{equation}
where each training sample
$
(Z_0,z^\text{src},\bP)
$
consist of $N$ encoded target views
$Z_0 = \{\mathcal{E}(x^i)\}_{i=1}^N$,
the encoded source view
$z^\text{src} = \mathcal{E}(x^\text{src})$
and the viewpoints $\bP$.
The network is conditioned on the source view and cameras and estimates the noise for all target views jointly.
Following~\cite{liu2023zero}, we also concatenate the input image code $z^\text{src}$ to $z_t$ along the channel dimension and add the CLIP~\cite{radford2021learning} encoding.

\begin{figure}
\centering
\includegraphics[width=\linewidth]{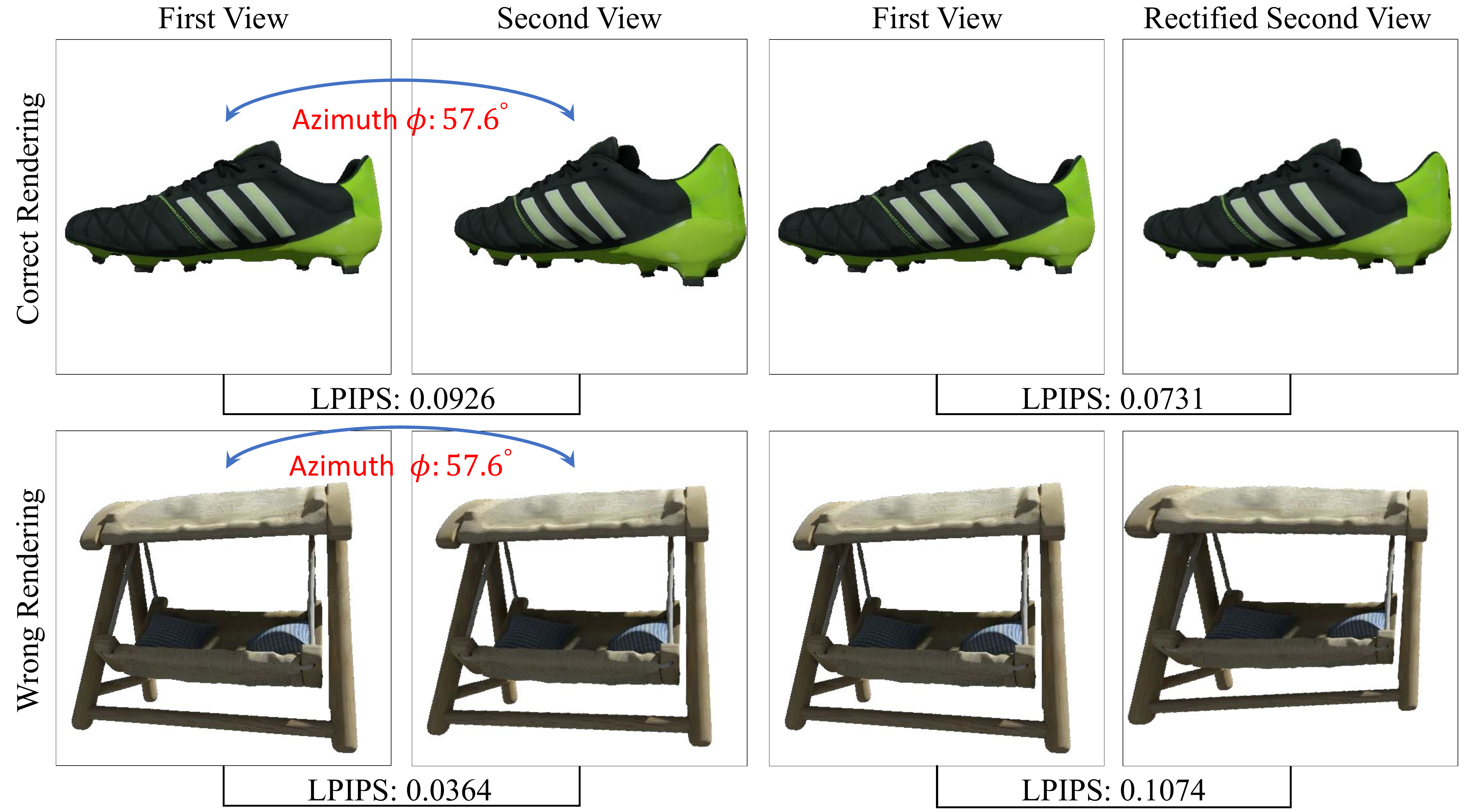}
\vspace{-1.5em}
\caption{\textbf{Perceptual Path Length Consistency (PPLC).}
To partly compensate for the viewpoint change, the second image is rectified w.r.t.~the first before comparison.
To illustrate the importance of using rectification, the figure shows two objects in a large azimuth $\phi:57.6^\circ$.
The top row shows to the left an ideally-rendered image pair, which however attains a large LPIPS score due to the view change.
To the right, rectification reduces this score.
The bottom row shows the opposite, where a pair of incorrectly rendered views has its LPIPS increased by rectification.
}%
\label{fig:pplc_rect}
\end{figure}

\subsection{Perceptual Path Length Consistency}%
\label{sec:PLC}

To quantify the consistency across different views, recent methods~\cite{watson2022novel,liu2023zero,weng2023consistent123,yang2023consistnet,voleti2024sv3d} trained 3D models from the sampled views and evaluated them on the remaining views.
However, this requires training a 3D model for each test instance, which is unfeasible for large-scale testing.
Here, we propose to use instead the pairwise perceptual distance~\cite{zhang2018unreasonable} to measure the consistency between generated views.
In particular, we first subdivide the 360$^\circ$ rendering path into linear segments and then calculate the LPIPS score between two neighboring generated images $x^i$ and $x^{i+1}$.
Naturally, these images differ due to the different viewpoints (\cref{fig:pplc_rect}).
We partly compensate for the viewpoint change by rectifying~\cite{hartley2003multiple} the second image w.r.t.~the first.
We then define Perceptual Path Length Consistency (PPLC) of a rendered sequence $\{x^i\}_{i=1}^N$ as follows:
\begin{equation}
\label{eq:pplc}
l_\text{pplc}
=
\mathbb{E}
\left[
    \frac{1}{\phi^2}\lVert
        \mathcal{F}(\text{Rect}(x^i)) -
        \mathcal{F}(\text{Rect}(x^{i+1}))
    \rVert_2^2
\right],
\end{equation}
where $\phi$ is the degree between views $x^i$ and $x^{i+1}$, which is set as $\phi=2\pi/50$, with the azimuth $7.2^\circ$ in all our video rendering.
$\mathcal{F}$ is a pre-trained network to ensure the metric matches with human perceptual similarity judgment.

\section{Experiments}%
\label{sec:exp}

\begin{figure*}[tb!]
\centering
\includegraphics[width=\linewidth]{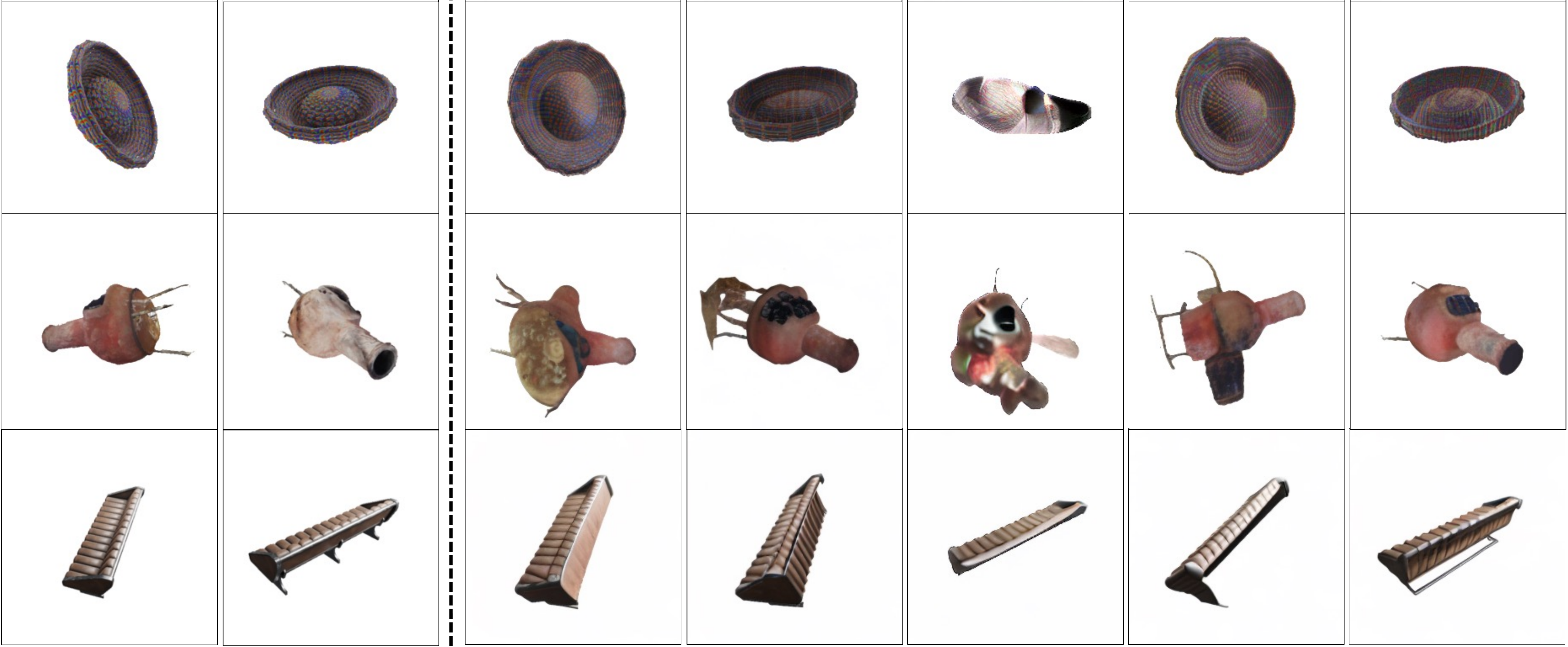}
\begin{picture}(0,0)
\put(-234,4){\footnotesize \textbf{(a)} Input View}
\put(-166,4){\footnotesize \textbf{(b)} Target View}
\put(-95,4){\footnotesize \textbf{(c)} Zero1-to-3~\cite{liu2023zero}}
\put(-29,4){\footnotesize \textbf{(d)} Zero123-XL~\cite{objaverseXL}}
\put(40,4){\footnotesize \textbf{(e)} SyncDreamer~\cite{liu2023syncdreamer}}
\put(111,4){\footnotesize \textbf{(f)} Consistent123~\cite{weng2023consistent123}}
\put(188,4){\footnotesize \textbf{(g)} Ours \method}
\end{picture}
\vspace{-6pt}
\caption{\textbf{Qualitative comparisons} on \texttt{Objaverse}.
Given a target pose, our \method significantly improves the accuracy of the generated pose compared to existing state-of-the-art methods.
Note that Zero123-XL~\cite{objaverseXL} is trained on the much larger \texttt{Objaverse-XL} dataset~\cite{objaverseXL}, which contains 10 million 3D objects.
More comparisons are provided in the supplement \cref{fig:sota_obj_app1,fig:sota_obj_app2}.}%
\label{fig:sota_obj}
\end{figure*}
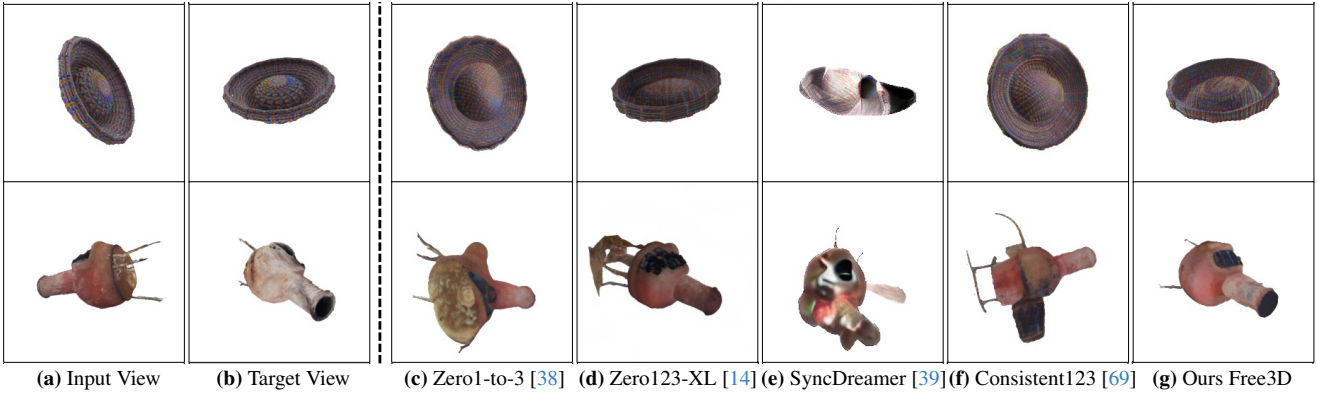
\begin{table}[tb!]
\centering
\setlength\tabcolsep{3pt}
\begin{tabular}{@{}l ccccc@{}}
\toprule
\multirow{2}{*}{\textbf{Method}} & \multicolumn{5}{c}{\textbf{Objaverse}\cite{deitke2023objaverse}} \\
\cline{2-6}
                                          & SSIM$\uparrow$  & LPIPS$\downarrow$ & FID$\downarrow$ & PPLC$\downarrow$ & Time$\downarrow$ \\ \midrule
Zero-1-to-3\cite{liu2023zero}                & 0.8462          & 0.0938            & 1.52            &          18.84        & \textbf{3s/44s}           \\
Zero123-XL\cite{objaverseXL}              & 0.8339          & 0.1098            & 1.67            &          25.61        & \textbf{3s/44s}           \\
SyncDreamer\cite{liu2023syncdreamer}      & 0.8063          & 0.1910            & 7.57            &          16.32        & 25s/77s          \\
Consistent123\cite{weng2023consistent123} & 0.8530          & 0.0913            & 1.48            &          17.89        & 4s/63s           \\ \midrule
Ours (\method)                            & \textbf{0.8620} & \textbf{0.0784}   & \textbf{1.21}   &         \textbf{10.82}         & 3s/52s           \\
\bottomrule
\end{tabular}
\caption{\textbf{Comparison with SoTA methods} on all 7,729 objects in \texttt{Objaverse test-set}.
Recent works like~\cite{liu2023zero,objaverseXL,weng2023consistent123} were originally evaluated using a subset of this data only due to the cost of training additional 3D models.
Unlike them, we directly evaluate the model on whole \texttt{test-set} without using additional 3D networks.
The inference time is for rendering a single target view and a $360^\circ$ video, respectively.}%
\label{tab:main_sota}
\end{table}


\subsection{Experimental Details}

\paragraph{Datasets.}

For fairness, our model is trained using the exact same protocol as Zero-1-to-3~\cite{liu2023zero}.
They render multiple views for 772,870 objects from \texttt{Objaverse} dataset~\cite{deitke2023objaverse}.
We use the identical \texttt{test} split as they do.
To assess how well our mode generalizes to other datasets, and how it compares to other models, we consider two more datasets:
\texttt{OmniObject3D}~\cite{wu2023omniobject3d} and
Google Scanned Objects (\texttt{GSO})~\cite{downs2022google}, which contain real-life scanned objects.
Since we do not use these datasets for training at all, we use the \emph{entirety} of \texttt{OmniObject3D} and \texttt{GSO} objects for evaluation (6,000 and 1,030 objects, respectively).

\paragraph{Metrics.}

We follow~\cite{liu2023zero,weng2023consistent123,liu2023syncdreamer} and assess the NVS quality by comparing the generated images and the ground-truth views at different levels of granularity, including PSNR, SSIM, LPIPS~\cite{zhang2018unreasonable}%
\footnote{\href{https://github.com/richzhang/PerceptualSimilarity}{https://github.com/richzhang/PerceptualSimilarity} ``\emph{squeeze}''-net.},
and Fr\'{e}chet Inception Distance (FID)~\cite{heusel2017gans}\footnote{\href{https://github.com/GaParmar/clean-fid}{https://github.com/GaParmar/clean-fid} ``\emph{clip\_vit\_b\_32}''-net.}.
Some of these metrics look at the statistics of individually reconstructed images rather than exact reconstructions, which is the correct approach given that the NVS task is inherently ambiguous; however, for the same reason, they are \emph{not suitable} for assessing multi-view consistency.
Instead, we use PPLC (\cref{sec:PLC}) to evaluate consistency.
To do so, we generate 50-frame videos along a pre-defined circular camera trajectory for each object and then compute the PPLC score between neighbouring frames.

\paragraph{Baselines.}

We compare \method to the state-of-the-art 
Zero-1-to-3~\cite{liu2023zero} 
and the follow-ups
Zero123-XL~\cite{objaverseXL} 
SyncDreamer~\cite{liu2023syncdreamer} 
and
Consistent123~\cite{weng2023consistent123}. 
While noticed other concurrent works in arXiv for NVS~\cite{ye2023consistent,kant2023invs,yang2023consistnet}, but, as no codes are available during the submission, we do not re-train for comparison.

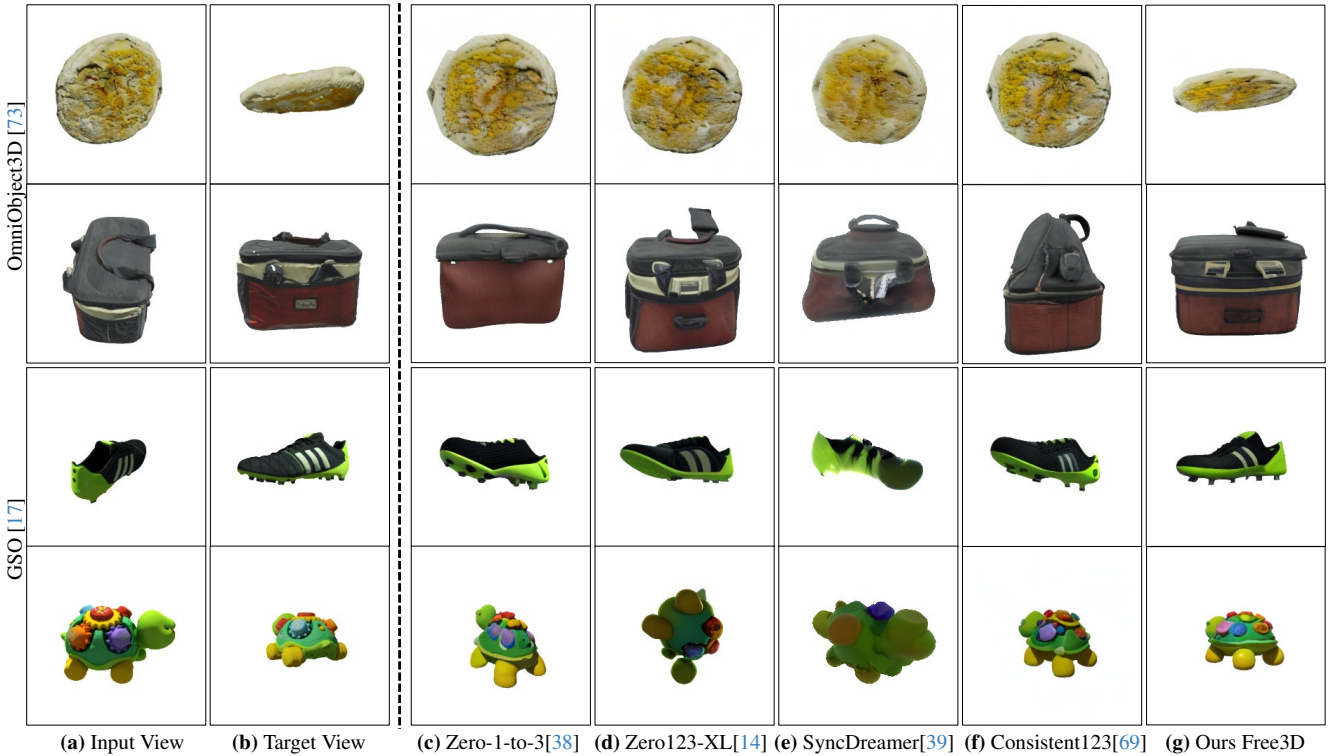
\begin{figure*}[tb!]
\centering
\includegraphics[width=\linewidth]{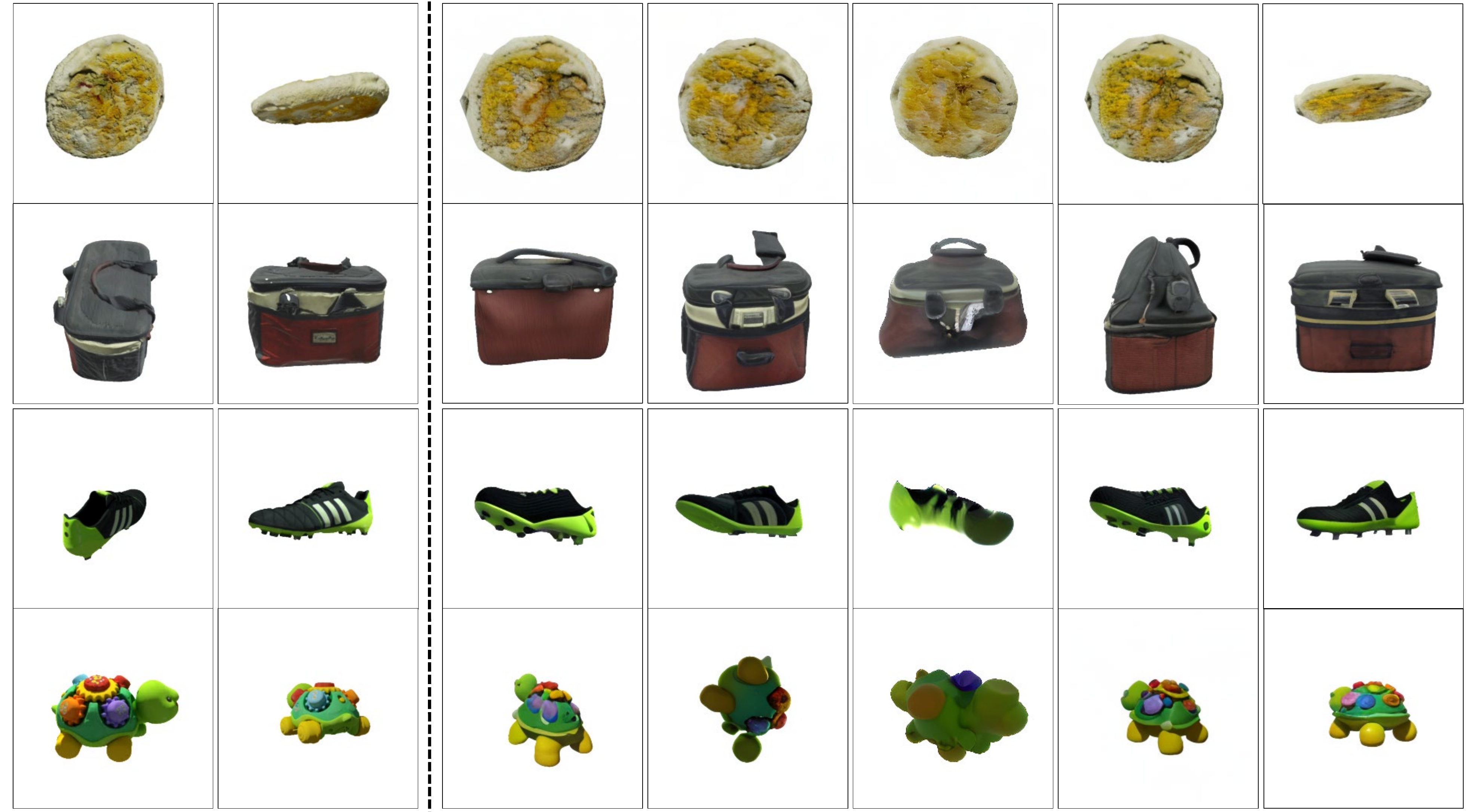}
\begin{picture}(0,0)
\put(-250,185){\rotatebox{90}{\footnotesize OmniObject3D~\cite{wu2023omniobject3d}}}
\put(-250,68){\rotatebox{90}{\footnotesize GSO~\cite{downs2022google}}}
\put(-230,4){\footnotesize \textbf{(a)} Input View}
\put(-165,4){\footnotesize \textbf{(b)} Target View}
\put(-95,4){\footnotesize \textbf{(c)} Zero-1-to-3\cite{liu2023zero}}
\put(-28,4){\footnotesize \textbf{(d)} Zero123-XL\cite{objaverseXL}}
\put(40,4){\footnotesize \textbf{(e)} SyncDreamer\cite{liu2023syncdreamer}}
\put(112,4){\footnotesize \textbf{(f)} Consistent123\cite{weng2023consistent123}}
\put(188,4){\footnotesize \textbf{(g)} Ours \method}
\end{picture}
\vspace{-6pt}
\caption{\textbf{Qualitative comparisons} on \texttt{OmniObject3D} (top two rows) and \texttt{GSO} (bottom two rows)
dataset.
Interestingly, exciting methods cannot deal with unconventional objects, such as the ``pie'' in the first row, while our \method is still robust for such a challenging scenario.
More comparisons are provided in supplemental \cref{fig:oo3d_gso_app1,fig:oo3d_gso_app2}.
}\label{fig:oo3d_gso}
\end{figure*}
\begin{table*}[tb!]
\centering
\setlength\tabcolsep{6pt}
\begin{tabular}{@{}l cccccc ccccc@{}}
\toprule
\multirow{2}{*}{\textbf{Method}}  & \multicolumn{5}{c}{\textbf{OmniObject3D}~\cite{wu2023omniobject3d}} && \multicolumn{5}{c}{\textbf{GSO}~\cite{downs2022google}} \\
\cline{2-6}\cline{8-12}
& PSNR$\uparrow$ & SSIM$\uparrow$ & LPIPS$\downarrow$ & FID$\downarrow$ & PPLC$\downarrow$ && PSNR$\uparrow$ & SSIM$\uparrow$ & LPIPS$\downarrow$ & FID$\downarrow$ & PPLC$\downarrow$ \\
\midrule
    Zero-1-to-3~\cite{liu2023zero}  & 16.84 & 0.7813 & 0.1321 & 1.73 & 24.58 && 19.65 & 0.8501 & 0.0758 & 3.24 & 33.15 \\
    Zero123-XL~\cite{objaverseXL}  & 17.11 & 0.7818 & 0.1291 & 1.51 & 21.33 &&  20.43 & 0.8589 & 0.0706 & 3.23 & 28.03 \\
    SyncDreamer~\cite{liu2023syncdreamer}  & 17.00 & 0.7941 & 0.1442 & 6.58 & 11.49 && 14.72 & 0.7835 & 0.1533 & 8.65  & 9.42 \\
    Consistent123~\cite{weng2023consistent123}$^*$  & 17.13 & 0.7821 & 0.1255 & 1.55 & 18.02 && 20.11 & 0.8553 & 0.0716 & 3.24 &  20.08 \\
\midrule
    Ours (\method) & \textbf{18.23} & \textbf{0.8090} & \textbf{0.0996} & \textbf{1.34} & \textbf{8.67} && \textbf{21.13} & \textbf{0.8686} & \textbf{0.0619} & \textbf{2.85} &  \textbf{9.10}\\
\bottomrule
\end{tabular}
\caption{\textbf{Generalizable results on unseen datasets}, including \texttt{OmniObject3D}~\cite{wu2023omniobject3d} and \texttt{GSO}~\cite{downs2022google}, with 6,000 and 1,030 3D instances, respectively.
Note that, although Zero123-XL~\cite{objaverseXL} is trained on a larger dataset, and shows better generalizability, the proposed \method still significantly outperforms it with \emph{precise} pose estimation for target views.}%
\label{tab:general_sota}
\end{table*}

\subsection{Assessing Quality}%
\label{s:quality}

\paragraph{Quantitative comparison.}

We first evaluate \method and other methods on the \texttt{Objaverse} dataset~\cite{deitke2023objaverse,liu2023zero}, where the model is trained.
Unlike previous work~\cite{liu2023zero,weng2023consistent123,liu2023syncdreamer} which consider only a subset of the test data due to expensive post-processing, we use the \emph{entire test set} of 7,729 3D objects.
Quantitative results in \cref{tab:main_sota} show that \method outperforms state-of-the-art models.
This includes Zero123-XL~\cite{objaverseXL} and SyncDreamer~\cite{liu2023syncdreamer}, which are trained on a much larger dataset and with explicit 3D volume representation, respectively.
While the concurrent Consistent123~\cite{weng2023consistent123} also utilizes a form of multi-view diffusion with cross-view attention, our \method significantly improves the quantitative results ($16\%$ relative improvement on LPIPS) on the large evaluation dataset.
This suggests that our RCN layer is the primary reason for the observed improvements.

\begin{figure*}[tb!]
\centering
\includegraphics[width=\linewidth]{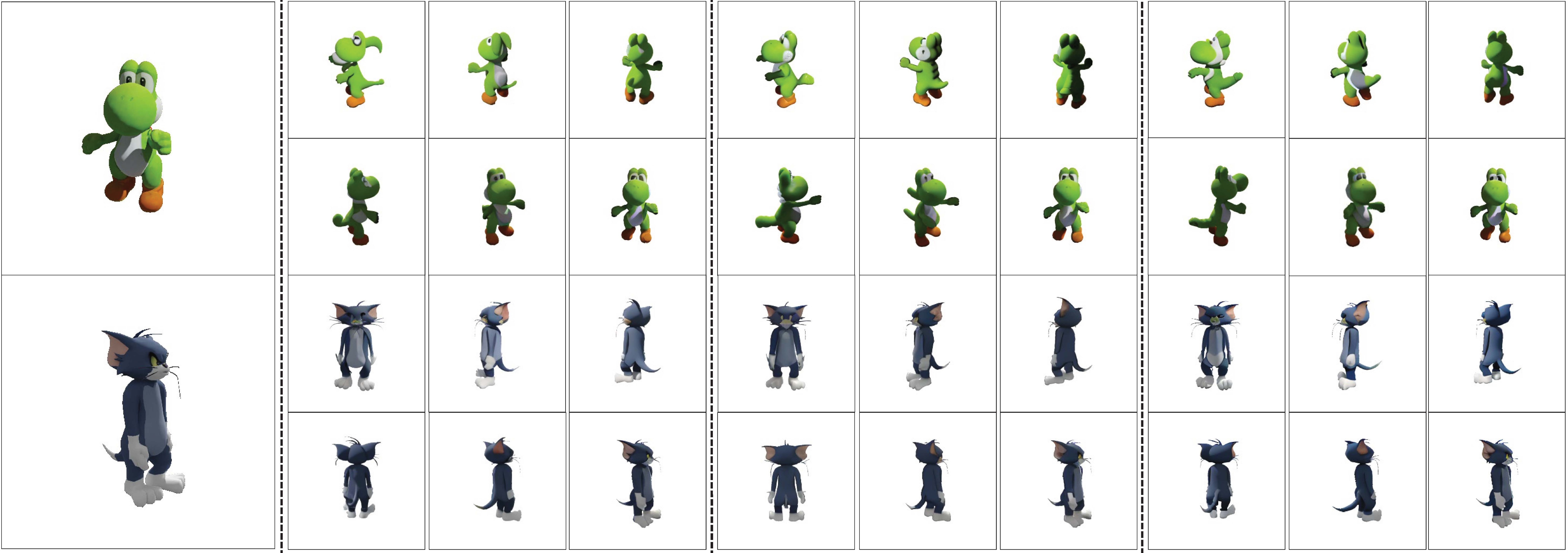}
\begin{picture}(0,0)
\put(-225,4){\footnotesize \textbf{(a)} Input View}
\put(-121,4){\footnotesize \textbf{(b)} Zero-1-to-3~\cite{liu2023zero}}
\put(10,4){\footnotesize \textbf{(c)} Consistent123~\cite{weng2023consistent123}}
\put(156,4){\footnotesize \textbf{(d)} Ours \method}
\end{picture}
\vspace{-0.5em}
\caption{\textbf{Qualitative new-view synthesis comparisons.} 
Zero-1-to-3~\cite{liu2023zero} generates diverse details (\eg various ears and tails) for different views in one sampling.
Consistent123~\cite{weng2023consistent123} improves it through the multi-view diffusion with the cross-view attention. 
However, it still requires training additional NeRF for 3D reconstruction.
For more comparisons, see the supplemental videos for better visualisation.}%
\label{fig:consistency}
\end{figure*}
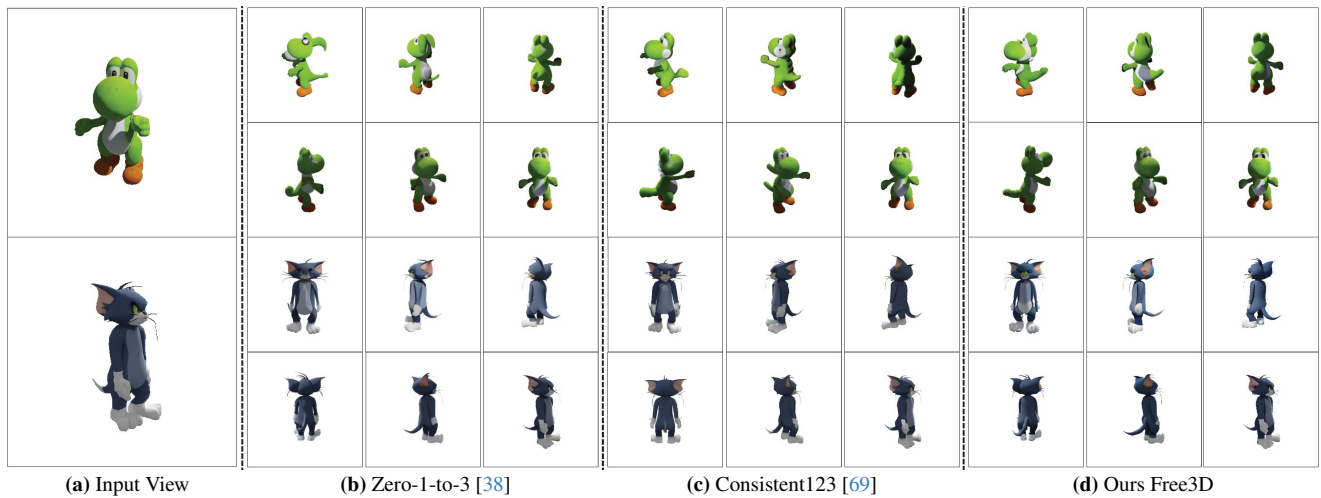
\begin{table*}
\centering
\setlength\tabcolsep{4pt}
\begin{tabular}{@{}ll cccccc ccccc@{}}
\bottomrule
& \multirow{2}{*}{\textbf{Method}} & \multicolumn{5}{c}{\textbf{Objaverse}~\cite{deitke2023objaverse}} && \multicolumn{5}{c}{\textbf{GSO}~\cite{downs2022google}} \\
\cline{3-7}\cline{9-13}
& & PSNR$\uparrow$ & SSIM$\uparrow$ & LPIPS$\downarrow$ & FID$\downarrow$ & PPLC$\downarrow$ && PSNR$\uparrow$ & SSIM$\uparrow$ & LPIPS$\downarrow$ & FID$\downarrow$ & PPLC$\downarrow$ \\
\midrule
$\mathbb{A}$ & Baseline Zero-1-to-3~\cite{liu2023zero} & 19.65 & 0.8462 & 0.0938 & 1.52 & 22.10 && 19.65 & 0.8501 & 0.0758 & 3.24 & 33.15 \\
$\mathbb{B}$ & + Input Ray Embeddings & 20.21 & 0.8550 & 0.0858 & 1.26 & 16.63 && 20.49 & 0.8617 & 0.0677 & 3.01 & 22.08 \\
$\mathbb{C}$ & + Multi-Scale Ray Emb. & 20.56 & 0.8609 & 0.0797 & 1.30 & 15.94 && 20.50 & 0.8615 & 0.0667 & 2.98 & 20.09 \\
$\mathbb{D}$ & + RCN & 20.78 & \textbf{0.8620} & 0.0784 & \textbf{1.21} & 15.67 && 21.13 & 0.8686 & 0.0619 & \textbf{2.85} & 18.48\\
$\mathbb{E}$ & + Pseudo-3D attention &  \textbf{20.81} & \textbf{0.8620} & \textbf{0.0781} & 1.25 & 14.76 && \textbf{21.20} &  \textbf{0.8697} & \textbf{0.0617} & 2.86 & 17.39 \\
$\mathbb{F}$ & $\mathbb{E}$ + noise sharing & --- & --- & --- & --- & \textbf{11.39} && --- & --- & --- & --- & \textbf{9.10} \\
\bottomrule
\end{tabular}
\caption{\textbf{Ablations} of \method design choices.
Here, to reduce the computational cost and testing time, we only evaluate 2,000 instances in \texttt{Objaverse} dataset, instead of running the whole dataset with 7,729 for 50-frame $360^\circ$ video rendering.
Therefore, the PPLC score is slightly different from the values reported in \cref{tab:main_sota}.
}%
\label{tab:config_abltion}
\vspace{-0.4cm}
\end{table*}
\paragraph{Qualitative comparison.}

Qualitative results
are visualized in \cref{fig:sota_obj}.
\method achieves better results even under challenging viewpoints.
SyncDreamer~\cite{liu2023syncdreamer} uses an \emph{explicit volumetric representation} representation, but can only generate views with fixed elevation ($30^\circ$), leading to worse results on synthesizing arbitrary target views.
While Consistent123~\cite{weng2023consistent123} aims to improve rendering consistency with a version of multi-view attention, they cannot directly improve pose accuracy.
The Zero123-XL~\cite{objaverseXL} learns to capture pose better than the baseline~\cite{liu2023zero} by training on the larger \texttt{Objaverse-XL}, but the pose is still \emph{not} very accurate.
Because of the RCN layer, \method shows no such pose errors and results in better images.

\subsection{Assessing Generalization}%
\label{s:generalization}

In \cref{tab:general_sota} and \cref{fig:oo3d_gso}, we validate the ability of \method to generalize to datasets not seen during training.
This includes the \texttt{OmniObject3D} and the \texttt{GSO} datasets, with 6,000 scanned objects in 190 categories and 1,030 scanned objects in 17 categories, respectively.
For this result, we directly test all trained models without any fine-tuning.

\paragraph{Quantitative comparison.}

In \cref{tab:general_sota}, \method
, which uses the same \texttt{training set} as Zero-1-to-3~\cite{liu2023zero}, 
outperforms the baseline and all state-of-the-art variants, including Zero123-XL~\cite{objaverseXL}, which is trained on a larger 3D dataset, and SyncDreamer~\cite{liu2023syncdreamer}, which utilizes a heavier 3D volumetric representation.
SyncDreamer improves the baseline on \texttt{OmniObject3D}, which has small elevation views change, while achieving worse results on \texttt{GSO} with random elevation angles.
Consistent123~\cite{weng2023consistent123} is also trained multiple views jointly.
However, their relative improvement is limited.
\Cref{tab:general_sota} shows that \method achieves very substantial improvements on all instantiations on both \texttt{OmniObject3D} and \texttt{GSO} datasets.
This further indicates that ray conditioning can successfully improve the pose accuracy of the target view.

\paragraph{Qualitative comparison.}

A qualitative comparison is given in \cref{fig:oo3d_gso} (more in supplemental \cref{fig:oo3d_gso_app1,fig:oo3d_gso_app2}).
Though \method is only trained on \texttt{objaverse} dataset,
it works quite well in the open-set setting.
Moreover, it shows significantly better results than all state-of-the-art models.

\paragraph{Results on real case.}

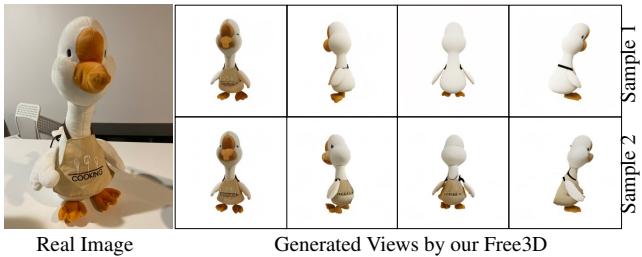
\begin{figure}[tb!]
    \centering
    \includegraphics[width=\linewidth]{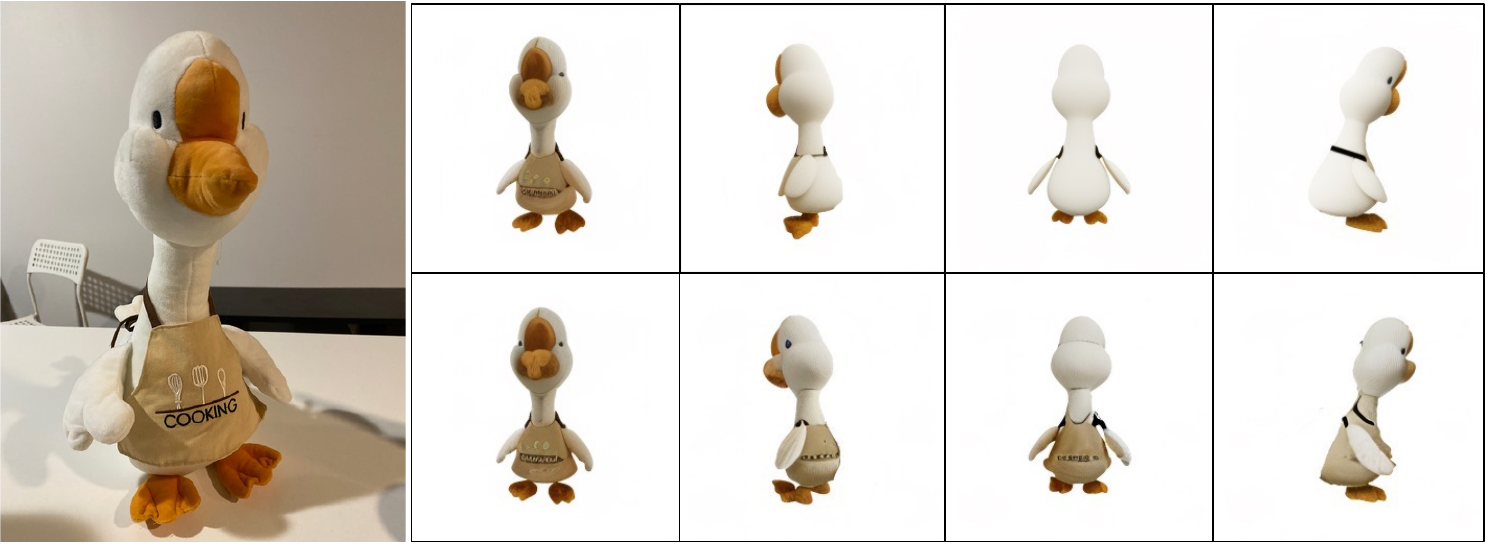}
    \begin{picture}(0,0)
    \put(-105,4){\footnotesize Real Image}
    \put(-15,4){\footnotesize Generated Views by our \method}
    \put(116,60){\rotatebox{90}{\footnotesize Sample 1}}
    \put(116,18){\rotatebox{90}{\footnotesize Sample 2}}
    \end{picture}
    \vspace{-0.5em}
    \caption{\textbf{Diverse NVS in real scenes.}
    {\color{red} Zoom in to see the details.}}%
    \label{fig:diverse}
\end{figure}

To further demonstrate the generalization of our \method to real images,
following~\cite{liu2023one2345++},
we run Segment Anything~\cite{kirillov2023segment} to extract the segments and show qualitative results in \cref{fig:teaser}.
Besides, we also show diverse results in \cref{fig:diverse} by sampling different noises.

\subsection{Assessing 3D consistency}

To assess 3D consistency, we render a $360^\circ$ video with the fixed elevation angle $\theta:=0^\circ$ and $50$ azimuth angles uniformly sampled in $[0^\circ, 360^\circ]$.
The quantitative results are reported in \cref{tab:main_sota,tab:general_sota} using the proposed PPLC score, and the qualitative results are shown in \cref{fig:teaser,fig:consistency}.
Consistent123~\cite{weng2023consistent123} still requires training additional NeRF to achieve 3D consistent video,
while the directly rendered videos are still flickering.
SyncDreamer~\cite{liu2023syncdreamer} obviously improves view consistency by using an explicit 3D volume representation.
However, it is trained only on a fixed elevation angle and can generate only 16 fixed frames for a video in its code.
Interestingly, \method, by using \emph{multi-view attention} and \emph{multi-view noise sharing}, is nearly as effective (while being much cheaper).
This is also partly due to ray conditioning, which can capture more precise poses of the target view and thus reduce ambiguity.
The additional results are provided as videos in the supplemental materials.

\subsection{Ablation Study}

We run a number of ablations to analyse \method.
Results are shown in \cref{tab:config_abltion} and discussed in detail next.

Our baseline configuration (denoted $\mathbb{A}$) is the same as Zero-1-to-3~\cite{liu2023zero}, which is derived from the SD model~\cite{rombach2022high} by replacing text conditioning with source image and target pose conditioning.
Then, in $\mathbb{B}$ we extend this model with ray conditioning by concatenating the ray embeddings to the source image $x^\text{src}$.
This alone improves the performance dramatically in both \texttt{Objaverse} and \texttt{GSO}.
In $\mathbb{C}$ we test injecting ray conditioning into each level of the diffusion UNet $\epsilon_\theta$, but the generalizable performance remains similar to $\mathbb{B}$ in the unseen \texttt{GSO} dataset.
In $\mathbb{D}$, we test instead the RCN layer, which results in further improvements on not only \texttt{Objaverse} dataset, but also on the new \texttt{GSO}.
In $\mathbb{E}$, we add \emph{multi-view attention} to exchange information between different frames.
As expected, this does \emph{not} improve the metrics that measure the quality of individual views, but it slightly improves the PPLC score, measuring consistency.
In $\mathbb{F}$, we further add \emph{multi-view noise sharing}, which significantly enhances the consistency between different views.
while preserving the quality of single-view rendering.
Rendered videos are provided in the supplement.

\section{Conclusion}%
\label{sec:conclusions}

We have introduced \method, an open-set single-view NVS method with state-of-the-art performance on various categories, yet bypass the requirement of building on heavy 3D representation or training additional auxiliary 3D models.
It is a simple approach that
(i) obtains data prior from an off-the-shelf pre-trained 2D image generator,
(ii) injects ray conditioning utilizing the new RCN layer to accurately code for the target pose,
and (iii) combines that with multi-view attention and noise sharing to improve multi-view consistency.
Experimental results show that \method significantly outperforms recent and concurrent state-of-the-art NVS models without incurring the cost of utilizing a 3D representation.
We hope that \method will serve as a new strong baseline for single-image NVS and inspire future research in this area.

\paragraph{Acknowledgements.}

This research is supported by ERC-CoG UNION 101001212.
Many thanks to Stanislaw Szymanowicz, Edgar Sucar, and Luke Melas-Kyriazi of VGG for insightful discussions and Ruining Li, Eldar Insafutdinov, and Yash Bhalgat of VGG for their helpful feedback.
We would also like to thank the authors of Zero-1-to-3~\cite{liu2023zero} and Objaverse-XL~\cite{objaverseXL} for their helpful discussions.

\paragraph{Ethics.}

We use the Objaverse~\cite{deitke2023objaverse}, OmniObject3D~\cite{wu2023omniobject3d}, and GSO~\cite{downs2022google} following their terms and conditions.
These datasets contain synthetic or scanned 3D objects, but, as far as we could determine, no personal data.
For further details on ethics, data protection, and copyright please see \url{https://www.robots.ox.ac.uk/~vedaldi/research/union/ethics.html}.

{
\small
\bibliographystyle{ieeenat_fullname}
\bibliography{main}

\begin{thebibliography}{82}
\providecommand{\natexlab}[1]{#1}
\providecommand{\url}[1]{\texttt{#1}}
\expandafter\ifx\csname urlstyle\endcsname\relax
  \providecommand{\doi}[1]{doi: #1}\else
  \providecommand{\doi}{doi: \begingroup \urlstyle{rm}\Url}\fi

\bibitem[Adelson and Wang(1992)]{adelson1992single}
Edward~H Adelson and John~YA Wang.
\newblock Single lens stereo with a plenoptic camera.
\newblock \emph{IEEE transactions on pattern analysis and machine intelligence (TPAMI)}, 14\penalty0 (2):\penalty0 99--106, 1992.

\bibitem[Adelson et~al.(1991)Adelson, Bergen, et~al.]{adelson1991plenoptic}
Edward~H Adelson, James~R Bergen, et~al.
\newblock The plenoptic function and the elements of early vision.
\newblock \emph{Computational models of visual processing}, 1\penalty0 (2):\penalty0 3--20, 1991.

\bibitem[Anciukevi{\v{c}}ius et~al.(2023)Anciukevi{\v{c}}ius, Xu, Fisher, Henderson, Bilen, Mitra, and Guerrero]{anciukevivcius2023renderdiffusion}
Titas Anciukevi{\v{c}}ius, Zexiang Xu, Matthew Fisher, Paul Henderson, Hakan Bilen, Niloy~J Mitra, and Paul Guerrero.
\newblock Renderdiffusion: Image diffusion for 3d reconstruction, inpainting and generation.
\newblock In \emph{Proceedings of the IEEE/CVF Conference on Computer Vision and Pattern Recognition (CVPR)}, pages 12608--12618, 2023.

\bibitem[Barron et~al.(2021)Barron, Mildenhall, Tancik, Hedman, Martin-Brualla, and Srinivasan]{barron2021mip}
Jonathan~T Barron, Ben Mildenhall, Matthew Tancik, Peter Hedman, Ricardo Martin-Brualla, and Pratul~P Srinivasan.
\newblock Mip-nerf: A multiscale representation for anti-aliasing neural radiance fields.
\newblock In \emph{Proceedings of the IEEE/CVF International Conference on Computer Vision (ICCV)}, pages 5855--5864, 2021.

\bibitem[Barron et~al.(2022)Barron, Mildenhall, Verbin, Srinivasan, and Hedman]{barron2022mip}
Jonathan~T Barron, Ben Mildenhall, Dor Verbin, Pratul~P Srinivasan, and Peter Hedman.
\newblock Mip-nerf 360: Unbounded anti-aliased neural radiance fields.
\newblock In \emph{Proceedings of the IEEE/CVF Conference on Computer Vision and Pattern Recognition (CVPR)}, pages 5470--5479, 2022.

\bibitem[Blattmann et~al.(2023{\natexlab{a}})Blattmann, Dockhorn, Kulal, Mendelevitch, Kilian, Lorenz, Levi, English, Voleti, Letts, et~al.]{blattmann2023stable}
Andreas Blattmann, Tim Dockhorn, Sumith Kulal, Daniel Mendelevitch, Maciej Kilian, Dominik Lorenz, Yam Levi, Zion English, Vikram Voleti, Adam Letts, et~al.
\newblock Stable video diffusion: Scaling latent video diffusion models to large datasets.
\newblock \emph{arXiv preprint arXiv:2311.15127}, 2023{\natexlab{a}}.

\bibitem[Blattmann et~al.(2023{\natexlab{b}})Blattmann, Rombach, Ling, Dockhorn, Kim, Fidler, and Kreis]{blattmann2023align}
Andreas Blattmann, Robin Rombach, Huan Ling, Tim Dockhorn, Seung~Wook Kim, Sanja Fidler, and Karsten Kreis.
\newblock Align your latents: High-resolution video synthesis with latent diffusion models.
\newblock In \emph{Proceedings of the IEEE/CVF Conference on Computer Vision and Pattern Recognition (CVPR)}, pages 22563--22575, 2023{\natexlab{b}}.

\bibitem[Chan et~al.(2023)Chan, Nagano, Chan, Bergman, Park, Levy, Aittala, Mello, Karras, and Wetzstein]{chan2023genvs}
Eric~R. Chan, Koki Nagano, Matthew~A. Chan, Alexander~W. Bergman, Jeong~Joon Park, Axel Levy, Miika Aittala, Shalini~De Mello, Tero Karras, and Gordon Wetzstein.
\newblock {GeNVS}: Generative novel view synthesis with {3D}-aware diffusion models.
\newblock In \emph{Proceedings of the International Conference on Computer Vision (ICCV)}, 2023.

\bibitem[Chen et~al.(2022)Chen, Xu, Geiger, Yu, and Su]{chen2022tensorf}
Anpei Chen, Zexiang Xu, Andreas Geiger, Jingyi Yu, and Hao Su.
\newblock Tensorf: Tensorial radiance fields.
\newblock In \emph{European Conference on Computer Vision (ECCV)}, pages 333--350. Springer, 2022.

\bibitem[Chen et~al.(2023{\natexlab{a}})Chen, Holalkere, Yan, Zhang, and Davis]{chen2023:ray-conditioning}
Eric~Ming Chen, Sidhanth Holalkere, Ruyu Yan, Kai Zhang, and Abe Davis.
\newblock Ray conditioning: Trading photo-realism for photo-consistency in multi-view image generation.
\newblock In \emph{Proceedings of the IEEE/CVF International Conference on Computer Vision (ICCV)}, 2023{\natexlab{a}}.

\bibitem[Chen and Williams(1993)]{chen1993view}
Shenchang~Eric Chen and Lance Williams.
\newblock View interpolation for image synthesis.
\newblock In \emph{Proceedings of the 20th Annual Conference on Computer Graphics and Interactive Techniques (SIGGRAPH)}, page 279–288, New York, NY, USA, 1993. Association for Computing Machinery.

\bibitem[Chen et~al.(2023{\natexlab{b}})Chen, Xu, Wu, Zheng, Cham, and Cai]{chen2023explicit}
Yuedong Chen, Haofei Xu, Qianyi Wu, Chuanxia Zheng, Tat-Jen Cham, and Jianfei Cai.
\newblock Explicit correspondence matching for generalizable neural radiance fields.
\newblock \emph{arXiv preprint arXiv:2304.12294}, 2023{\natexlab{b}}.

\bibitem[Debevec et~al.(1996)Debevec, Taylor, and Malik]{debec1996modeling}
Paul~E Debevec, Camillo~J Taylor, and Jitendra Malik.
\newblock Modeling and rendering architecture from photographs: A hybrid geometry-and image-based approach.
\newblock In \emph{Proceedings of the 23th Annual Conference on Computer Graphics and Interactive Techniques (SIGGRAPH)}, 1996.

\bibitem[Deitke et~al.(2023{\natexlab{a}})Deitke, Liu, Wallingford, Ngo, Michel, Kusupati, Fan, Laforte, Voleti, Gadre, VanderBilt, Kembhavi, Vondrick, Gkioxari, Ehsani, Schmidt, and Farhadi]{objaverseXL}
Matt Deitke, Ruoshi Liu, Matthew Wallingford, Huong Ngo, Oscar Michel, Aditya Kusupati, Alan Fan, Christian Laforte, Vikram Voleti, Samir~Yitzhak Gadre, Eli VanderBilt, Aniruddha Kembhavi, Carl Vondrick, Georgia Gkioxari, Kiana Ehsani, Ludwig Schmidt, and Ali Farhadi.
\newblock Objaverse-xl: A universe of 10m+ 3d objects.
\newblock \emph{Advances in Neural Information Processing Systems (NeurIPS)}, 2023{\natexlab{a}}.

\bibitem[Deitke et~al.(2023{\natexlab{b}})Deitke, Schwenk, Salvador, Weihs, Michel, VanderBilt, Schmidt, Ehsani, Kembhavi, and Farhadi]{deitke2023objaverse}
Matt Deitke, Dustin Schwenk, Jordi Salvador, Luca Weihs, Oscar Michel, Eli VanderBilt, Ludwig Schmidt, Kiana Ehsani, Aniruddha Kembhavi, and Ali Farhadi.
\newblock Objaverse: A universe of annotated 3d objects.
\newblock In \emph{Proceedings of the IEEE/CVF Conference on Computer Vision and Pattern Recognition (CVPR)}, pages 13142--13153, 2023{\natexlab{b}}.

\bibitem[Dhariwal and Nichol(2021)]{dhariwal2021diffusion}
Prafulla Dhariwal and Alexander Nichol.
\newblock Diffusion models beat gans on image synthesis.
\newblock \emph{Advances in neural information processing systems (NeurIPS)}, 34:\penalty0 8780--8794, 2021.

\bibitem[Downs et~al.(2022)Downs, Francis, Koenig, Kinman, Hickman, Reymann, McHugh, and Vanhoucke]{downs2022google}
Laura Downs, Anthony Francis, Nate Koenig, Brandon Kinman, Ryan Hickman, Krista Reymann, Thomas~B McHugh, and Vincent Vanhoucke.
\newblock Google scanned objects: A high-quality dataset of 3d scanned household items.
\newblock In \emph{2022 International Conference on Robotics and Automation (ICRA)}, pages 2553--2560. IEEE, 2022.

\bibitem[Dumoulin et~al.(2016)Dumoulin, Shlens, and Kudlur]{dumoulin2016learned}
Vincent Dumoulin, Jonathon Shlens, and Manjunath Kudlur.
\newblock A learned representation for artistic style.
\newblock In \emph{International Conference on Learning Representations (ICLR)}, 2016.

\bibitem[Fridovich-Keil et~al.(2022)Fridovich-Keil, Yu, Tancik, Chen, Recht, and Kanazawa]{fridovich2022plenoxels}
Sara Fridovich-Keil, Alex Yu, Matthew Tancik, Qinhong Chen, Benjamin Recht, and Angjoo Kanazawa.
\newblock Plenoxels: Radiance fields without neural networks.
\newblock In \emph{Proceedings of the IEEE/CVF Conference on Computer Vision and Pattern Recognition (CVPR)}, pages 5501--5510, 2022.

\bibitem[Goodfellow et~al.(2014)Goodfellow, Pouget-Abadie, Mirza, Xu, Warde-Farley, Ozair, Courville, and Bengio]{goodfellow2014generative}
Ian Goodfellow, Jean Pouget-Abadie, Mehdi Mirza, Bing Xu, David Warde-Farley, Sherjil Ozair, Aaron Courville, and Yoshua Bengio.
\newblock Generative adversarial nets.
\newblock \emph{Advances in neural information processing systems}, 27, 2014.

\bibitem[Guo et~al.(2023)Guo, Yang, Rao, Wang, Qiao, Lin, and Dai]{guo2023animatediff}
Yuwei Guo, Ceyuan Yang, Anyi Rao, Yaohui Wang, Yu Qiao, Dahua Lin, and Bo Dai.
\newblock Animatediff: Animate your personalized text-to-image diffusion models without specific tuning.
\newblock \emph{arXiv preprint arXiv:2307.04725}, 2023.

\bibitem[Hartley and Zisserman(2003)]{hartley2003multiple}
Richard Hartley and Andrew Zisserman.
\newblock \emph{Multiple view geometry in computer vision}.
\newblock Cambridge university press, 2003.

\bibitem[Heusel et~al.(2017)Heusel, Ramsauer, Unterthiner, Nessler, and Hochreiter]{heusel2017gans}
Martin Heusel, Hubert Ramsauer, Thomas Unterthiner, Bernhard Nessler, and Sepp Hochreiter.
\newblock Gans trained by a two time-scale update rule converge to a local nash equilibrium.
\newblock In \emph{Proceedings of the 31st International Conference on Neural Information Processing Systems (NeurIPS)}, pages 6626--6637, 2017.

\bibitem[Ho et~al.(2020)Ho, Jain, and Abbeel]{ho2020denoising}
Jonathan Ho, Ajay Jain, and Pieter Abbeel.
\newblock Denoising diffusion probabilistic models.
\newblock \emph{Advances in neural information processing systems (NeurIPS)}, 33:\penalty0 6840--6851, 2020.

\bibitem[Ho et~al.(2022)Ho, Chan, Saharia, Whang, Gao, Gritsenko, Kingma, Poole, Norouzi, Fleet, et~al.]{ho2022imagen}
Jonathan Ho, William Chan, Chitwan Saharia, Jay Whang, Ruiqi Gao, Alexey Gritsenko, Diederik~P Kingma, Ben Poole, Mohammad Norouzi, David~J Fleet, et~al.
\newblock Imagen video: High definition video generation with diffusion models.
\newblock \emph{arXiv preprint arXiv:2210.02303}, 2022.

\bibitem[Huang and Belongie(2017)]{huang2017arbitrary}
Xun Huang and Serge Belongie.
\newblock Arbitrary style transfer in real-time with adaptive instance normalization.
\newblock In \emph{Proceedings of the IEEE international conference on computer vision (ICCV)}, pages 1501--1510, 2017.

\bibitem[Huang et~al.(2022)Huang, Stojanov, Thai, Jampani, and Rehg]{huang2022planes}
Zixuan Huang, Stefan Stojanov, Anh Thai, Varun Jampani, and James~M Rehg.
\newblock Planes vs. chairs: Category-guided 3d shape learning without any 3d cues.
\newblock In \emph{European Conference on Computer Vision}, pages 727--744. Springer, 2022.

\bibitem[Jiang et~al.(2023)Jiang, Tang, Chang, Song, Wang, and Cao]{jiang2023efficient}
Yifan Jiang, Hao Tang, Jen-Hao~Rick Chang, Liangchen Song, Zhangyang Wang, and Liangliang Cao.
\newblock Efficient-3dim: Learning a generalizable single-image novel-view synthesizer in one day.
\newblock \emph{arXiv preprint arXiv:2310.03015}, 2023.

\bibitem[Kanazawa et~al.(2018)Kanazawa, Tulsiani, Efros, and Malik]{kanazawa2018learning}
Angjoo Kanazawa, Shubham Tulsiani, Alexei~A Efros, and Jitendra Malik.
\newblock Learning category-specific mesh reconstruction from image collections.
\newblock In \emph{Proceedings of the European Conference on Computer Vision (ECCV)}, pages 371--386, 2018.

\bibitem[Kant et~al.(2023)Kant, Siarohin, Vasilkovsky, Guler, Ren, Tulyakov, and Gilitschenski]{kant2023invs}
Yash Kant, Aliaksandr Siarohin, Michael Vasilkovsky, Riza~Alp Guler, Jian Ren, Sergey Tulyakov, and Igor Gilitschenski.
\newblock invs: Repurposing diffusion inpainters for novel view synthesis.
\newblock \emph{arXiv preprint arXiv:2310.16167}, 2023.

\bibitem[Karnewar et~al.(2023)Karnewar, Vedaldi, Novotny, and Mitra]{karnewar2023holodiffusion}
Animesh Karnewar, Andrea Vedaldi, David Novotny, and Niloy~J Mitra.
\newblock Holodiffusion: Training a 3d diffusion model using 2d images.
\newblock In \emph{Proceedings of the IEEE/CVF Conference on Computer Vision and Pattern Recognition (CVPR)}, pages 18423--18433, 2023.

\bibitem[Karras et~al.(2019)Karras, Laine, and Aila]{karras2019style}
Tero Karras, Samuli Laine, and Timo Aila.
\newblock A style-based generator architecture for generative adversarial networks.
\newblock In \emph{Proceedings of the IEEE/CVF conference on computer vision and pattern recognition}, pages 4401--4410, 2019.

\bibitem[Kerbl et~al.(2023)Kerbl, Kopanas, Leimk{\"u}hler, and Drettakis]{kerbl20233d}
Bernhard Kerbl, Georgios Kopanas, Thomas Leimk{\"u}hler, and George Drettakis.
\newblock 3d gaussian splatting for real-time radiance field rendering.
\newblock \emph{ACM Transactions on Graphics (ToG)}, 42\penalty0 (4):\penalty0 1--14, 2023.

\bibitem[Kingma and Welling(2014)]{kingma2013auto}
Diederik~P Kingma and Max Welling.
\newblock Auto-encoding variational bayes.
\newblock In \emph{Proceedings of the International Conference on Learning Representations (ICLR)}, 2014.

\bibitem[Kirillov et~al.(2023)Kirillov, Mintun, Ravi, Mao, Rolland, Gustafson, Xiao, Whitehead, Berg, Lo, et~al.]{kirillov2023segment}
Alexander Kirillov, Eric Mintun, Nikhila Ravi, Hanzi Mao, Chloe Rolland, Laura Gustafson, Tete Xiao, Spencer Whitehead, Alexander~C Berg, Wan-Yen Lo, et~al.
\newblock Segment anything.
\newblock In \emph{Proceedings of the IEEE/CVF International Conference on Computer Vision (ICCV)}, pages 4015--4026, 2023.

\bibitem[Liu et~al.(2023{\natexlab{a}})Liu, Shi, Chen, Zhang, Xu, Wei, Chen, Zeng, Gu, and Su]{liu2023one2345++}
Minghua Liu, Ruoxi Shi, Linghao Chen, Zhuoyang Zhang, Chao Xu, Xinyue Wei, Hansheng Chen, Chong Zeng, Jiayuan Gu, and Hao Su.
\newblock One-2-3-45++: Fast single image to 3d objects with consistent multi-view generation and 3d diffusion.
\newblock \emph{arXiv preprint arXiv:2311.07885}, 2023{\natexlab{a}}.

\bibitem[Liu et~al.(2023{\natexlab{b}})Liu, Xu, Jin, Chen, Xu, Su, et~al.]{liu2023one}
Minghua Liu, Chao Xu, Haian Jin, Linghao Chen, Zexiang Xu, Hao Su, et~al.
\newblock One-2-3-45: Any single image to 3d mesh in 45 seconds without per-shape optimization.
\newblock \emph{arXiv preprint arXiv:2306.16928}, 2023{\natexlab{b}}.

\bibitem[Liu et~al.(2023{\natexlab{c}})Liu, Wu, Van~Hoorick, Tokmakov, Zakharov, and Vondrick]{liu2023zero}
Ruoshi Liu, Rundi Wu, Basile Van~Hoorick, Pavel Tokmakov, Sergey Zakharov, and Carl Vondrick.
\newblock Zero-1-to-3: Zero-shot one image to 3d object.
\newblock In \emph{Proceedings of the IEEE/CVF International Conference on Computer Vision (ICCV)}, pages 9298--9309, 2023{\natexlab{c}}.

\bibitem[Liu et~al.(2023{\natexlab{d}})Liu, Lin, Zeng, Long, Liu, Komura, and Wang]{liu2023syncdreamer}
Yuan Liu, Cheng Lin, Zijiao Zeng, Xiaoxiao Long, Lingjie Liu, Taku Komura, and Wenping Wang.
\newblock Syncdreamer: Learning to generate multiview-consistent images from a single-view image.
\newblock \emph{arXiv preprint arXiv:2309.03453}, 2023{\natexlab{d}}.

\bibitem[Lombardi et~al.(2019)Lombardi, Simon, Saragih, Schwartz, Lehrmann, and Sheikh]{lombardi2019neural}
Stephen Lombardi, Tomas Simon, Jason Saragih, Gabriel Schwartz, Andreas Lehrmann, and Yaser Sheikh.
\newblock Neural volumes: learning dynamic renderable volumes from images.
\newblock \emph{ACM Transactions on Graphics (TOG)}, 38\penalty0 (4):\penalty0 1--14, 2019.

\bibitem[Melas-Kyriazi et~al.(2023{\natexlab{a}})Melas-Kyriazi, Laina, Rupprecht, and Vedaldi]{melas2023realfusion}
Luke Melas-Kyriazi, Iro Laina, Christian Rupprecht, and Andrea Vedaldi.
\newblock Realfusion: 360deg reconstruction of any object from a single image.
\newblock In \emph{Proceedings of the IEEE/CVF Conference on Computer Vision and Pattern Recognition (CVPR)}, pages 8446--8455, 2023{\natexlab{a}}.

\bibitem[Melas-Kyriazi et~al.(2023{\natexlab{b}})Melas-Kyriazi, Rupprecht, and Vedaldi]{melas2023pc2}
Luke Melas-Kyriazi, Christian Rupprecht, and Andrea Vedaldi.
\newblock Pc2: Projection-conditioned point cloud diffusion for single-image 3d reconstruction.
\newblock In \emph{Proceedings of the IEEE/CVF Conference on Computer Vision and Pattern Recognition (CVPR)}, pages 12923--12932, 2023{\natexlab{b}}.

\bibitem[Mildenhall et~al.(2020)Mildenhall, Srinivasan, Tancik, Barron, Ramamoorthi, and Ng]{mildenhall2020nerf}
B Mildenhall, PP Srinivasan, M Tancik, JT Barron, R Ramamoorthi, and R Ng.
\newblock Nerf: Representing scenes as neural radiance fields for view synthesis.
\newblock In \emph{Proceedings of the European conference on computer vision (ECCV)}, 2020.

\bibitem[M{\"u}ller et~al.(2022)M{\"u}ller, Evans, Schied, and Keller]{muller2022instant}
Thomas M{\"u}ller, Alex Evans, Christoph Schied, and Alexander Keller.
\newblock Instant neural graphics primitives with a multiresolution hash encoding.
\newblock \emph{ACM Transactions on Graphics (ToG)}, 41\penalty0 (4):\penalty0 1--15, 2022.

\bibitem[Park et~al.(2017)Park, Yang, Yumer, Ceylan, and Berg]{park2017transformation}
Eunbyung Park, Jimei Yang, Ersin Yumer, Duygu Ceylan, and Alexander~C Berg.
\newblock Transformation-grounded image generation network for novel 3d view synthesis.
\newblock In \emph{Proceedings of the ieee conference on computer vision and pattern recognition (CVPR)}, pages 3500--3509, 2017.

\bibitem[Park et~al.(2019)Park, Florence, Straub, Newcombe, and Lovegrove]{park2019deepsdf}
Jeong~Joon Park, Peter Florence, Julian Straub, Richard Newcombe, and Steven Lovegrove.
\newblock Deepsdf: Learning continuous signed distance functions for shape representation.
\newblock In \emph{Proceedings of the IEEE/CVF conference on computer vision and pattern recognition (CVPR)}, pages 165--174, 2019.

\bibitem[Poole et~al.(2023)Poole, Jain, Barron, and Mildenhall]{poole2023dreamfusion}
Ben Poole, Ajay Jain, Jonathan~T Barron, and Ben Mildenhall.
\newblock Dreamfusion: Text-to-3d using 2d diffusion.
\newblock In \emph{The Eleventh International Conference on Learning Representations (ICLR)}, 2023.

\bibitem[Purushwalkam and Naik(2023)]{purushwalkam2023conrad}
Senthil Purushwalkam and Nikhil Naik.
\newblock Conrad: Image constrained radiance fields for 3d generation from a single image.
\newblock \emph{Advances in Neural Information Processing Systems (NeurIPS)}, 2023.

\bibitem[Qian et~al.(2023)Qian, Mai, Hamdi, Ren, Siarohin, Li, Lee, Skorokhodov, Wonka, Tulyakov, et~al.]{qian2023magic123}
Guocheng Qian, Jinjie Mai, Abdullah Hamdi, Jian Ren, Aliaksandr Siarohin, Bing Li, Hsin-Ying Lee, Ivan Skorokhodov, Peter Wonka, Sergey Tulyakov, et~al.
\newblock Magic123: One image to high-quality 3d object generation using both 2d and 3d diffusion priors.
\newblock \emph{arXiv preprint arXiv:2306.17843}, 2023.

\bibitem[Radford et~al.(2021)Radford, Kim, Hallacy, Ramesh, Goh, Agarwal, Sastry, Askell, Mishkin, Clark, et~al.]{radford2021learning}
Alec Radford, Jong~Wook Kim, Chris Hallacy, Aditya Ramesh, Gabriel Goh, Sandhini Agarwal, Girish Sastry, Amanda Askell, Pamela Mishkin, Jack Clark, et~al.
\newblock Learning transferable visual models from natural language supervision.
\newblock In \emph{International conference on machine learning (ICML)}, pages 8748--8763. PMLR, 2021.

\bibitem[Rombach et~al.(2022)Rombach, Blattmann, Lorenz, Esser, and Ommer]{rombach2022high}
Robin Rombach, Andreas Blattmann, Dominik Lorenz, Patrick Esser, and Bj{\"o}rn Ommer.
\newblock High-resolution image synthesis with latent diffusion models.
\newblock In \emph{Proceedings of the IEEE/CVF conference on computer vision and pattern recognition (CVPR)}, pages 10684--10695, 2022.

\bibitem[Saharia et~al.(2022)Saharia, Chan, Saxena, Li, Whang, Denton, Ghasemipour, Gontijo~Lopes, Karagol~Ayan, Salimans, et~al.]{saharia2022photorealistic}
Chitwan Saharia, William Chan, Saurabh Saxena, Lala Li, Jay Whang, Emily~L Denton, Kamyar Ghasemipour, Raphael Gontijo~Lopes, Burcu Karagol~Ayan, Tim Salimans, et~al.
\newblock Photorealistic text-to-image diffusion models with deep language understanding.
\newblock \emph{Advances in Neural Information Processing Systems (NeurIPS)}, 35:\penalty0 36479--36494, 2022.

\bibitem[Sajjadi et~al.(2022{\natexlab{a}})Sajjadi, Duckworth, Mahendran, van Steenkiste, Pavetic, Lucic, Guibas, Greff, and Kipf]{sajjadi2022object}
Mehdi~SM Sajjadi, Daniel Duckworth, Aravindh Mahendran, Sjoerd van Steenkiste, Filip Pavetic, Mario Lucic, Leonidas~J Guibas, Klaus Greff, and Thomas Kipf.
\newblock Object scene representation transformer.
\newblock \emph{Advances in Neural Information Processing Systems (NeurIPS)}, 35:\penalty0 9512--9524, 2022{\natexlab{a}}.

\bibitem[Sajjadi et~al.(2022{\natexlab{b}})Sajjadi, Meyer, Pot, Bergmann, Greff, Radwan, Vora, Lu{\v{c}}i{\'c}, Duckworth, Dosovitskiy, et~al.]{sajjadi2022scene}
Mehdi~SM Sajjadi, Henning Meyer, Etienne Pot, Urs Bergmann, Klaus Greff, Noha Radwan, Suhani Vora, Mario Lu{\v{c}}i{\'c}, Daniel Duckworth, Alexey Dosovitskiy, et~al.
\newblock Scene representation transformer: Geometry-free novel view synthesis through set-latent scene representations.
\newblock In \emph{Proceedings of the IEEE/CVF Conference on Computer Vision and Pattern Recognition (CVPR)}, pages 6229--6238, 2022{\natexlab{b}}.

\bibitem[Sargent et~al.(2023)Sargent, Li, Shah, Herrmann, Yu, Zhang, Chan, Lagun, Fei-Fei, Sun, et~al.]{sargent2023zeronvs}
Kyle Sargent, Zizhang Li, Tanmay Shah, Charles Herrmann, Hong-Xing Yu, Yunzhi Zhang, Eric~Ryan Chan, Dmitry Lagun, Li Fei-Fei, Deqing Sun, et~al.
\newblock Zeronvs: Zero-shot 360-degree view synthesis from a single real image.
\newblock \emph{arXiv preprint arXiv:2310.17994}, 2023.

\bibitem[Schuhmann et~al.(2022)Schuhmann, Beaumont, Vencu, Gordon, Wightman, Cherti, Coombes, Katta, Mullis, Wortsman, et~al.]{schuhmann2022laion}
Christoph Schuhmann, Romain Beaumont, Richard Vencu, Cade Gordon, Ross Wightman, Mehdi Cherti, Theo Coombes, Aarush Katta, Clayton Mullis, Mitchell Wortsman, et~al.
\newblock Laion-5b: An open large-scale dataset for training next generation image-text models.
\newblock \emph{Advances in Neural Information Processing Systems (NeurIPS)}, 35:\penalty0 25278--25294, 2022.

\bibitem[Shi et~al.(2023)Shi, Wang, Ye, Long, Li, and Yang]{shi2023mvdream}
Yichun Shi, Peng Wang, Jianglong Ye, Mai Long, Kejie Li, and Xiao Yang.
\newblock Mvdream: Multi-view diffusion for 3d generation.
\newblock \emph{arXiv preprint arXiv:2308.16512}, 2023.

\bibitem[Singer et~al.(2022)Singer, Polyak, Hayes, Yin, An, Zhang, Hu, Yang, Ashual, Gafni, et~al.]{singer2022make}
Uriel Singer, Adam Polyak, Thomas Hayes, Xi Yin, Jie An, Songyang Zhang, Qiyuan Hu, Harry Yang, Oron Ashual, Oran Gafni, et~al.
\newblock Make-a-video: Text-to-video generation without text-video data.
\newblock In \emph{The Eleventh International Conference on Learning Representations (ICLR)}, 2022.

\bibitem[Sitzmann et~al.(2019)Sitzmann, Zollh{\"o}fer, and Wetzstein]{sitzmann2019scene}
Vincent Sitzmann, Michael Zollh{\"o}fer, and Gordon Wetzstein.
\newblock Scene representation networks: Continuous 3d-structure-aware neural scene representations.
\newblock \emph{Advances in Neural Information Processing Systems (NeurIPS)}, 32, 2019.

\bibitem[Sitzmann et~al.(2021)Sitzmann, Rezchikov, Freeman, Tenenbaum, and Durand]{sitzmann2021light}
Vincent Sitzmann, Semon Rezchikov, Bill Freeman, Josh Tenenbaum, and Fredo Durand.
\newblock Light field networks: Neural scene representations with single-evaluation rendering.
\newblock \emph{Advances in Neural Information Processing Systems (NeurIPS)}, 34:\penalty0 19313--19325, 2021.

\bibitem[Sohl-Dickstein et~al.(2015)Sohl-Dickstein, Weiss, Maheswaranathan, and Ganguli]{sohl2015deep}
Jascha Sohl-Dickstein, Eric Weiss, Niru Maheswaranathan, and Surya Ganguli.
\newblock Deep unsupervised learning using nonequilibrium thermodynamics.
\newblock In \emph{International conference on machine learning (ICML)}, pages 2256--2265. PMLR, 2015.

\bibitem[Suhail et~al.(2022)Suhail, Esteves, Sigal, and Makadia]{suhail2022light}
Mohammed Suhail, Carlos Esteves, Leonid Sigal, and Ameesh Makadia.
\newblock Light field neural rendering.
\newblock In \emph{Proceedings of the IEEE/CVF Conference on Computer Vision and Pattern Recognition (CVPR)}, pages 8269--8279, 2022.

\bibitem[Szymanowicz et~al.(2023)Szymanowicz, Rupprecht, and Vedaldi]{szymanowicz23viewset}
Stanislaw Szymanowicz, Christian Rupprecht, and Andrea Vedaldi.
\newblock Viewset diffusion: (0-)image-conditioned {3D} generative models from {2D} data.
\newblock In \emph{Proceedings of the International Conference on Computer Vision (ICCV)}, 2023.

\bibitem[Tang et~al.(2023)Tang, Wang, Zhang, Zhang, Yi, Ma, and Chen]{tang2023make}
Junshu Tang, Tengfei Wang, Bo Zhang, Ting Zhang, Ran Yi, Lizhuang Ma, and Dong Chen.
\newblock Make-it-3d: High-fidelity 3d creation from a single image with diffusion prior.
\newblock \emph{arXiv preprint arXiv:2303.14184}, 2023.

\bibitem[Tatarchenko et~al.(2016)Tatarchenko, Dosovitskiy, and Brox]{tatarchenko2016multi}
Maxim Tatarchenko, Alexey Dosovitskiy, and Thomas Brox.
\newblock Multi-view 3d models from single images with a convolutional network.
\newblock In \emph{Computer Vision--ECCV 2016: 14th European Conference, Amsterdam, The Netherlands, October 11--14, 2016, Proceedings, Part VII 14}, pages 322--337. Springer, 2016.

\bibitem[Tseng et~al.(2023)Tseng, Li, Kim, Alsisan, Huang, and Kopf]{tseng2023consistent}
Hung-Yu Tseng, Qinbo Li, Changil Kim, Suhib Alsisan, Jia-Bin Huang, and Johannes Kopf.
\newblock Consistent view synthesis with pose-guided diffusion models.
\newblock In \emph{Proceedings of the IEEE/CVF Conference on Computer Vision and Pattern Recognition (CVPR)}, pages 16773--16783, 2023.

\bibitem[Voleti et~al.(2024)Voleti, Yao, Boss, Letts, Pankratz, Tochilkin, Laforte, Rombach, and Jampani]{voleti2024sv3d}
Vikram Voleti, Chun-Han Yao, Mark Boss, Adam Letts, David Pankratz, Dmitry Tochilkin, Christian Laforte, Robin Rombach, and Varun Jampani.
\newblock Sv3d: Novel multi-view synthesis and 3d generation from a single image using latent video diffusion.
\newblock \emph{arXiv preprint arXiv:2403.12008}, 2024.

\bibitem[Watson et~al.(2023)Watson, Chan, Brualla, Ho, Tagliasacchi, and Norouzi]{watson2022novel}
Daniel Watson, William Chan, Ricardo~Martin Brualla, Jonathan Ho, Andrea Tagliasacchi, and Mohammad Norouzi.
\newblock Novel view synthesis with diffusion models.
\newblock In \emph{The Eleventh International Conference on Learning Representations (ICLR)}, 2023.

\bibitem[Weng et~al.(2023)Weng, Yang, Wang, Li, Zhang, Chen, and Zhang]{weng2023consistent123}
Haohan Weng, Tianyu Yang, Jianan Wang, Yu Li, Tong Zhang, CL Chen, and Lei Zhang.
\newblock Consistent123: Improve consistency for one image to 3d object synthesis.
\newblock \emph{arXiv preprint arXiv:2310.08092}, 2023.

\bibitem[Wiles et~al.(2020)Wiles, Gkioxari, Szeliski, and Johnson]{wiles2020synsin}
Olivia Wiles, Georgia Gkioxari, Richard Szeliski, and Justin Johnson.
\newblock Synsin: End-to-end view synthesis from a single image.
\newblock In \emph{Proceedings of the IEEE/CVF Conference on Computer Vision and Pattern Recognition}, pages 7467--7477, 2020.

\bibitem[Wu et~al.(2023{\natexlab{a}})Wu, Liu, Zhao, Kale, Bui, Yu, Lin, Zhang, and Chang]{wu2023uncovering}
Qiucheng Wu, Yujian Liu, Handong Zhao, Ajinkya Kale, Trung Bui, Tong Yu, Zhe Lin, Yang Zhang, and Shiyu Chang.
\newblock Uncovering the disentanglement capability in text-to-image diffusion models.
\newblock In \emph{Proceedings of the IEEE/CVF Conference on Computer Vision and Pattern Recognition (CVPR)}, pages 1900--1910, 2023{\natexlab{a}}.

\bibitem[Wu et~al.(2023{\natexlab{b}})Wu, Chen, Yang, Guo, Li, and Zhang]{wu2023lamp}
Ruiqi Wu, Liangyu Chen, Tong Yang, Chunle Guo, Chongyi Li, and Xiangyu Zhang.
\newblock Lamp: Learn a motion pattern for few-shot-based video generation.
\newblock \emph{arXiv preprint arXiv:2310.10769}, 2023{\natexlab{b}}.

\bibitem[Wu et~al.(2023{\natexlab{c}})Wu, Zhang, Fu, Wang, Ren, Pan, Wu, Yang, Wang, Qian, et~al.]{wu2023omniobject3d}
Tong Wu, Jiarui Zhang, Xiao Fu, Yuxin Wang, Jiawei Ren, Liang Pan, Wayne Wu, Lei Yang, Jiaqi Wang, Chen Qian, et~al.
\newblock Omniobject3d: Large-vocabulary 3d object dataset for realistic perception, reconstruction and generation.
\newblock In \emph{Proceedings of the IEEE/CVF Conference on Computer Vision and Pattern Recognition (CVPR)}, pages 803--814, 2023{\natexlab{c}}.

\bibitem[Xiong et~al.(2023)Xiong, Ma, Sun, Han, and Xie]{xiong2023light}
Yifeng Xiong, Haoyu Ma, Shanlin Sun, Kun Han, and Xiaohui Xie.
\newblock Light field diffusion for single-view novel view synthesis.
\newblock \emph{arXiv preprint arXiv:2309.11525}, 2023.

\bibitem[Xu et~al.(2024)Xu, Shi, Yifan, Chen, Yang, Peng, Shen, and Wetzstein]{xu2024grm}
Yinghao Xu, Zifan Shi, Wang Yifan, Hansheng Chen, Ceyuan Yang, Sida Peng, Yujun Shen, and Gordon Wetzstein.
\newblock Grm: Large gaussian reconstruction model for efficient 3d reconstruction and generation.
\newblock \emph{arXiv preprint arXiv:2403.14621}, 2024.

\bibitem[Yang et~al.(2023)Yang, Cheng, Duan, Ji, and Li]{yang2023consistnet}
Jiayu Yang, Ziang Cheng, Yunfei Duan, Pan Ji, and Hongdong Li.
\newblock Consistnet: Enforcing 3d consistency for multi-view images diffusion.
\newblock \emph{arXiv preprint arXiv:2310.10343}, 2023.

\bibitem[Ye et~al.(2024)Ye, Wang, Li, Shi, and Wang]{ye2023consistent}
Jianglong Ye, Peng Wang, Kejie Li, Yichun Shi, and Heng Wang.
\newblock Consistent-1-to-3: Consistent image to 3d view synthesis via geometry-aware diffusion models.
\newblock In \emph{Proceedings of the International Conference on 3D Vision (3DV)}, 2024.

\bibitem[Yu et~al.(2021)Yu, Ye, Tancik, and Kanazawa]{yu2021pixelnerf}
Alex Yu, Vickie Ye, Matthew Tancik, and Angjoo Kanazawa.
\newblock pixelnerf: Neural radiance fields from one or few images.
\newblock In \emph{Proceedings of the IEEE/CVF Conference on Computer Vision and Pattern Recognition (CVPR)}, pages 4578--4587, 2021.

\bibitem[Zhang et~al.(2020)Zhang, Riegler, Snavely, and Koltun]{zhang2020nerf++}
Kai Zhang, Gernot Riegler, Noah Snavely, and Vladlen Koltun.
\newblock Nerf++: Analyzing and improving neural radiance fields.
\newblock \emph{arXiv preprint arXiv:2010.07492}, 2020.

\bibitem[Zhang et~al.(2018)Zhang, Isola, Efros, Shechtman, and Wang]{zhang2018unreasonable}
Richard Zhang, Phillip Isola, Alexei~A Efros, Eli Shechtman, and Oliver Wang.
\newblock The unreasonable effectiveness of deep features as a perceptual metric.
\newblock In \emph{Proceedings of the IEEE Conference on Computer Vision and Pattern Recognition (CVPR)}, pages 586--595, 2018.

\bibitem[Zheng et~al.(2022)Zheng, Vuong, Cai, and Phung]{zheng2022movq}
Chuanxia Zheng, Tung-Long Vuong, Jianfei Cai, and Dinh Phung.
\newblock Movq: Modulating quantized vectors for high-fidelity image generation.
\newblock \emph{Advances in Neural Information Processing Systems (NeurIPS)}, 35:\penalty0 23412--23425, 2022.

\bibitem[Zhou and Tulsiani(2023)]{zhou2023sparsefusion}
Zhizhuo Zhou and Shubham Tulsiani.
\newblock Sparsefusion: Distilling view-conditioned diffusion for 3d reconstruction.
\newblock In \emph{Proceedings of the IEEE/CVF Conference on Computer Vision and Pattern Recognition (CVPR)}, pages 12588--12597, 2023.

\end{thebibliography}
}

\clearpage
\appendix\onecolumn
\renewcommand{\theequation}{\thesection.\arabic{equation}}
\setcounter{equation}{0}
\renewcommand{\thefigure}{\thesection.\arabic{figure}}
\setcounter{figure}{0}
\renewcommand{\thetable}{\thesection.\arabic{table}}
\setcounter{table}{0}
\setcounter{page}{1}
\maketitlesupplementary
\newpage

The supplementary materials are organized as follows:

\begin{itemize}[itemsep=0pt]
\item A video to illuminate our work and the rendered videos.

\item Introduction for the baseline diffusion model.

\item Experiment details.

\item Results for more single view NVS\@.
\end{itemize}

\section{Background: Diffusion Generators}%
\label{s:background}

In order to achieve sufficient generalization to operate in an \emph{open-set} category setting, \method builds on a pre-trained 2D image generation, and specifically Stable Diffusion (SD)~\cite{rombach2022high}.
SD is a Latent Diffusion Model (LDM) trained on billions of text-image pairs from LAION-5B~\cite{schuhmann2022laion}.
It consists of two stages.
The first stage
embeds the given image $x_0\in\mathbb{R}^{H\times W\times 3}$ in a latent space $z\in\mathbb{R}^{\frac{H}{f}\times \frac{H}{f}\times c}$ through an autoencoder $\mathcal{E}: x \mapsto z$, paired with a decoder $\mathcal{D}: z \mapsto x$, which reconstructs the image ($x = \mathcal{D} \circ \mathcal{E}(x)$).
The second stage uses diffusion to model the distribution $p(z|y)$ over such latent codes, where $y$ lumps any conditioning information (\eg, text, image, or viewpoint).
Diffusion involves a forward noising process that gradually perturbs the given latent $z_0=z$ by adding the Gaussian noise $\epsilon$ in a Markovian fashion:
\begin{equation}\label{eq:ldm_forward}
z_t = \sqrt{\bar{\alpha}_t}z_0 + \sqrt{1-\bar{\alpha}_t}\epsilon,
\quad
\epsilon\sim\mathcal{N}(0,I),
\end{equation}
producing a sequence $z_t$, $t=1,\dots,T$, and $\bar{\alpha}_t:=\prod_{s=1}^t\alpha_s$, $\alpha_t:=1-\beta_t$ denote the noise strength at different steps. 
$\{\beta_t\}_{t=1}^T$ is a pre-defined variance schedule. 
Ultimately, $p(z_T|y)$ is approximately normal;
we can thus easily sample $z_T$, and then go back to $z_0$ via the backward denoising process using the predicted noise:
\begin{equation}\label{eq:ldm_backward}
z_{t-1} =
\frac{1}{\sqrt{\alpha_t}}
\left(
    z_t -
    \frac{1-\alpha_t}{\sqrt{1-\bar{\alpha_t}}}
    \epsilon_\theta(z_t,t,y)
\right)
+ \sigma_t\epsilon,
\end{equation}
where $\epsilon_\theta$ is typically an UNet~\cite{dhariwal2021diffusion}, 
and $\{\sigma_t\}_{t=1}^T$ is another control of noise $\epsilon$,
which is also a pre-defined schedule corresponding to the schedule $\beta_t$ and introduces uncertainty for the synthesis of different views.
Similar to the vanilla DDPM~\cite{ho2020denoising}, SD uses the following training objective to optimize the UNet $\epsilon_\theta$:
\begin{equation}\label{eq:ldm_loss}
    \mathcal{L} = \mathbb{E}_{z_0, y, \epsilon\sim\mathcal{N}(0,I), t}
    \left[
    \lVert
        \epsilon - \epsilon_\theta(z_t, t, y)
    \rVert^2_2
    \right],
\end{equation}

\section{Experiment Details}

The Stable Diffusion (SD), originally trained for text-to-images generation,
requires adaptation to suit image-conditional NVS tasks.
Following Zero-1-to-3~\cite{liu2023zero}, we utilize the image-to-image Stable Diffusion checkpoints\footnote{\href{https://huggingface.co/spaces/lambdalabs/stable-diffusion-image-variations}{https://huggingface.co/spaces/lambdalabs/stable-diffusion-image-variations}}. Our baseline code is built upon the Zero-1-to-3~\cite{liu2023zero}\footnote{\href{https://github.com/cvlab-columbia/zero123}{https://github.com/cvlab-columbia/zero123}}.
Hyperparameters are configured in accordance with the default settings of the baseline code.
The \emph{ray conditioning normalisation (RCN)} is incorporated into each ResNet block within the diffusion Unet $\epsilon_\theta$, while the \emph{pseudo-3D cross-attention} is introduced after the original CLIP-conditional cross-attention layer (as illustrated in \cref{fig:framework}).

Instead of directly providing $\br_{uv}=(\bo\times\bd_{uv},\bd_{uv})$ to the network for modulating the features, 
we embed them into higher-dimensional features, following the approach of NeRF~\cite{mildenhall2020nerf} and LFN~\cite{sitzmann2021light}.
In particular, we employ the element-wise mapping $\br \mapsto [\br,\sin(f_1\pi\br),\cos(f_1\pi\br),\cdots,\sin(f_K\pi\br),\cos(f_K\pi\br)]$, where $K$ is the number of Fourier bands, and $f_k$ is equally spaced to the sampling rate.
In all experiments, $K$ is set as 6, leading to 
$78=2\times3\times K_{\bo} + 3 + 2\times3\times K_{\bd} + 3=(6+6)\times6+6$ dimensional features (as depicted in~\cref{fig:framework}(a)). 

\paragraph{Training Details.}
Our model was trained on 4$\times$ A40 48GB GPUs in two stages:
\textbf{i)} We first finetuned the model with RCN, utilizing a batch size of 256 for 3 days on random camera viewpoints, enhancing the pose accuracy for target views.
\textbf{ii)} Subsequently, the pseudo-3D cross-attention was finetuned on the 4 nearest views, employing a batch size of 192 for 2 days.
In the second stage, different views from one instance were perturbed by adding noise from the same time step $t$. 

In an alternative approach during the first stage, we initially attempted to jointly train the pseudo-3D cross-attention with random camera viewpoints.
However, the performance is worse than the configuration $\mathbb{D}$.
We believe this is because the camera viewpoints have a large gap along these random views in the rendered datasets, making it harder to calculate the similarity across these frames.
In all experiments, we use AdamW with a learning rate of $10^{-5}$ for the old parameters in the original diffusion Unet $\epsilon_\theta$ and a $10\times$ larger learning rate for new parameters, namely the parameters for RCN and Pseudo-3D cross-attention.

\paragraph{Inference Details.}
At the testing phase, we configure the diffusion model with a sampling step set to $T=50$.
The computational time for rendering a novel view using our proposed \method is approximately 3 seconds, utilizing an A6000 GPU.
For a fair comparison, all models are evaluated on the same A6000 GPU employing the same batch size of 4.
This batch size is chosen due to the operational constraints of syndreamer~\cite{liu2023syncdreamer}, which can only run such a small size.
Additionally, we also utilize the CFG with a scale $s=3$ to guide the rendering for each target view.

\paragraph{360$^\circ$ Video Rendering.}
To render a 360$^\circ$ video, we establish a circle trajectory by uniformly subdividing the azimuth $\phi$ into discrete intervals of $\frac{2\pi}{50} = 7.2^\circ$,
while the elevation $\theta$ and the distance $z$ remain fixed.
For each 3D instance, we replicate the same latent variable $z_T$ over 50 frames, which can minimize temporal flickering across different views.
Additionally, we also set the parameter $\sigma_t$ in \cref{eq:ldm_backward} to zero, thereby further mitigating uncertainty introduced by varying noise patterns.

\section{More Visual Results}

\begin{figure*}[tb!]
    \centering
    \includegraphics[width=\linewidth]{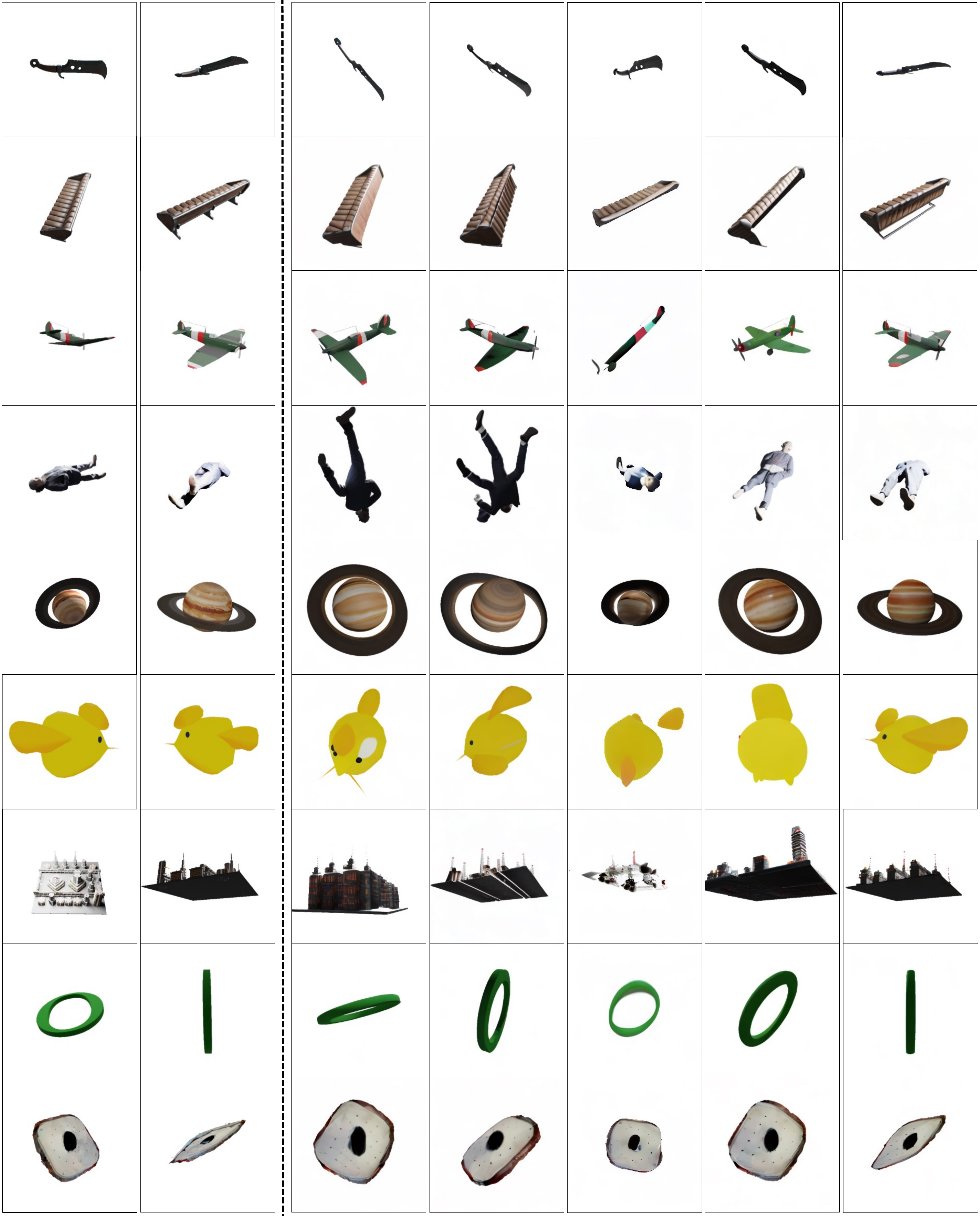}
    \begin{picture}(0,0)
    \put(-236,6){\footnotesize \textbf{(a)} Input View}
    \put(-165,6){\footnotesize \textbf{(b)} Target View}
    \put(-90,6){\footnotesize \textbf{(c)} Zero123\cite{liu2023zero}}
    \put(-28,6){\footnotesize \textbf{(d)} Zero123-XL\cite{objaverseXL}}
    \put(40,6){\footnotesize \textbf{(e)} SyncDreamer\cite{liu2023syncdreamer}}
    \put(112,6){\footnotesize \textbf{(f)} Consistent123\cite{weng2023consistent123}}
    \put(188,6){\footnotesize \textbf{(g)} Ours \method}
    \end{picture}
    \vspace{-10pt}
    \caption{\textbf{Qualitative comparisons on \texttt{Objaverse} dataset.}
    Given the exact target pose, the proposed \method significantly improves the pose precision compared to existing state-of-the-art methods.
    }
    \label{fig:sota_obj_app1}
\end{figure*}
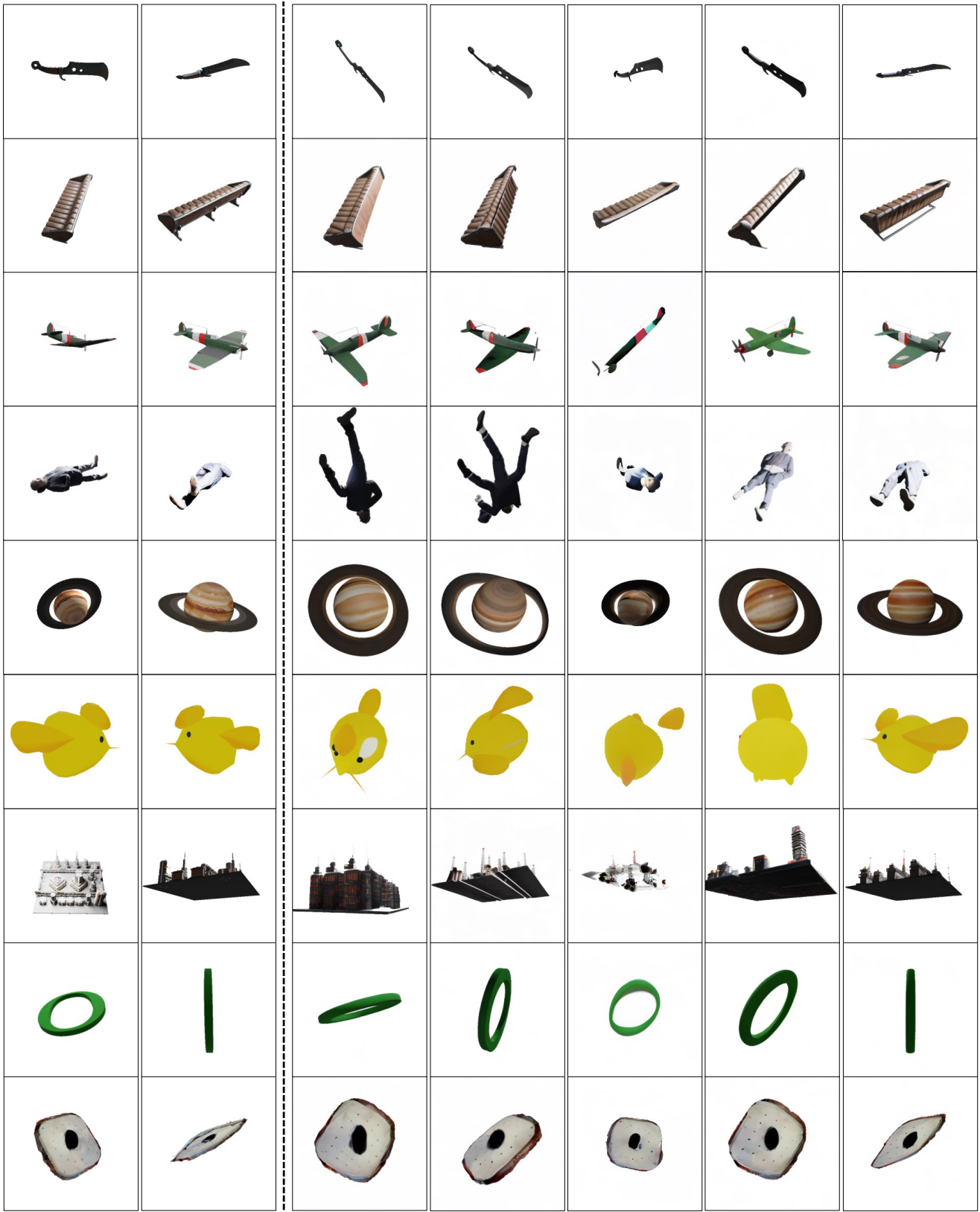

\begin{figure*}[tb!]
    \centering
    \includegraphics[width=\linewidth]{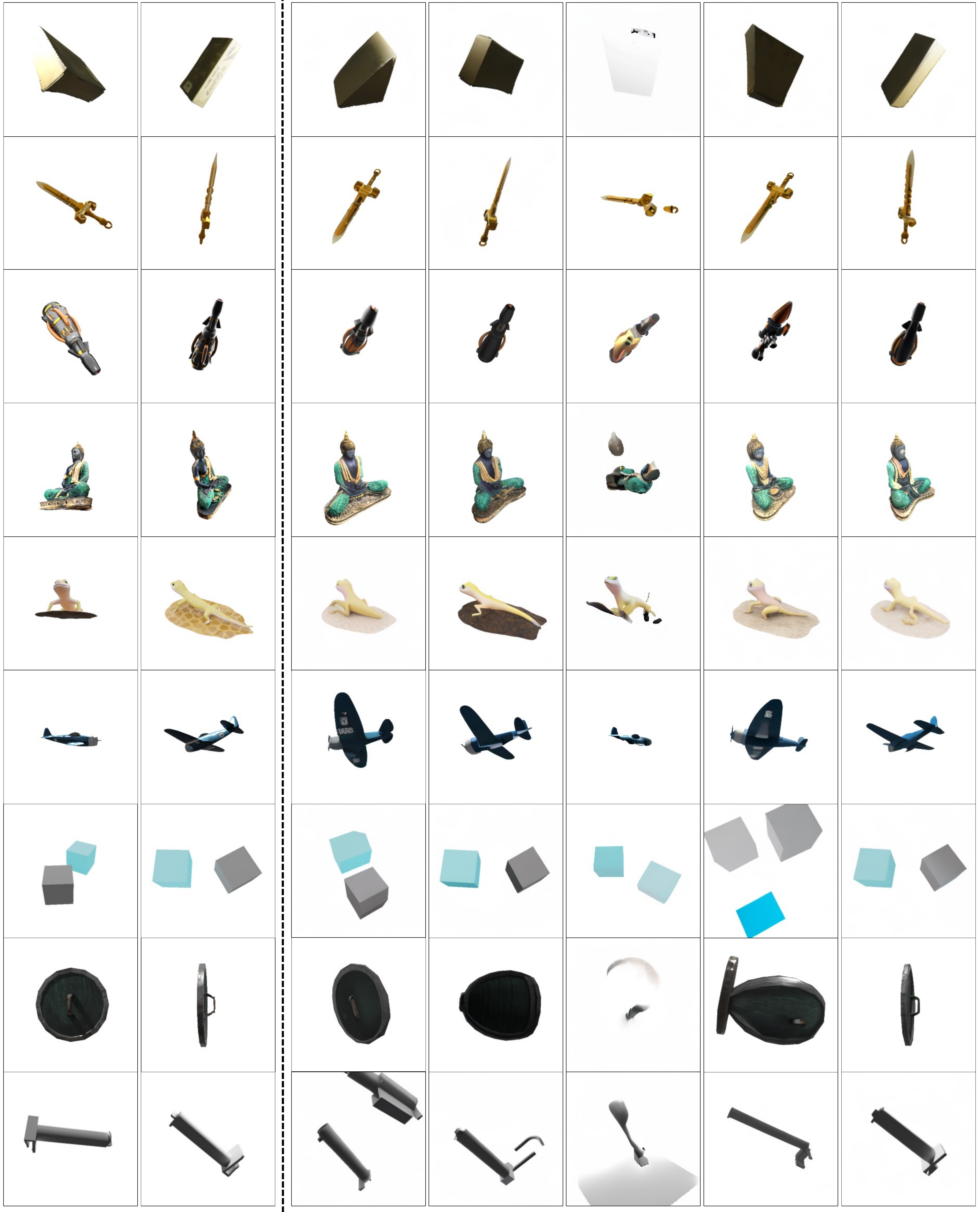}
    \begin{picture}(0,0)
    \put(-236,6){\footnotesize \textbf{(a)} Input View}
    \put(-165,6){\footnotesize \textbf{(b)} Target View}
    \put(-90,6){\footnotesize \textbf{(c)} Zero123\cite{liu2023zero}}
    \put(-28,6){\footnotesize \textbf{(d)} Zero123-XL\cite{objaverseXL}}
    \put(40,6){\footnotesize \textbf{(e)} SyncDreamer\cite{liu2023syncdreamer}}
    \put(112,6){\footnotesize \textbf{(f)} Consistent123\cite{weng2023consistent123}}
    \put(188,6){\footnotesize \textbf{(g)} Ours \method}
    \end{picture}
    \vspace{-10pt}
    \caption{\textbf{Qualitative comparisons on \texttt{Objaverse} dataset.}
    Given the exact target pose, the proposed \method significantly improves the pose precision compared to existing state-of-the-art methods.
    }
    \label{fig:sota_obj_app2}
\end{figure*}

\paragraph{More results on Objaverse NVS.} 
In \cref{fig:sota_obj_app1,fig:sota_obj_app2}, we present more visual comparisons on \texttt{Objaverse} datasets~\cite{deitke2023objaverse} that given one input image and the target viewpoint, all models render the target novel view.
This is an extension of \cref{fig:sota_obj} in the main paper.

Here, all examples shown come from the corresponding \texttt{test-set} following the split, as in Zero-1-to-3~\cite{liu2023zero}.
These examples are good evidence that our \method is suitable for \emph{open-set} categories NVS, where it can generate semantically reasonable content with visually realistic appearances across various categories.
More importantly, compared to existing state-of-the-art methods, the \method provides better results with a more precise pose for the target novel view.
This observation suggests that the RCN is able to provide better viewpoint perception for the NVS.

\begin{figure*}[tb!]
    \centering
    \includegraphics[width=\linewidth]{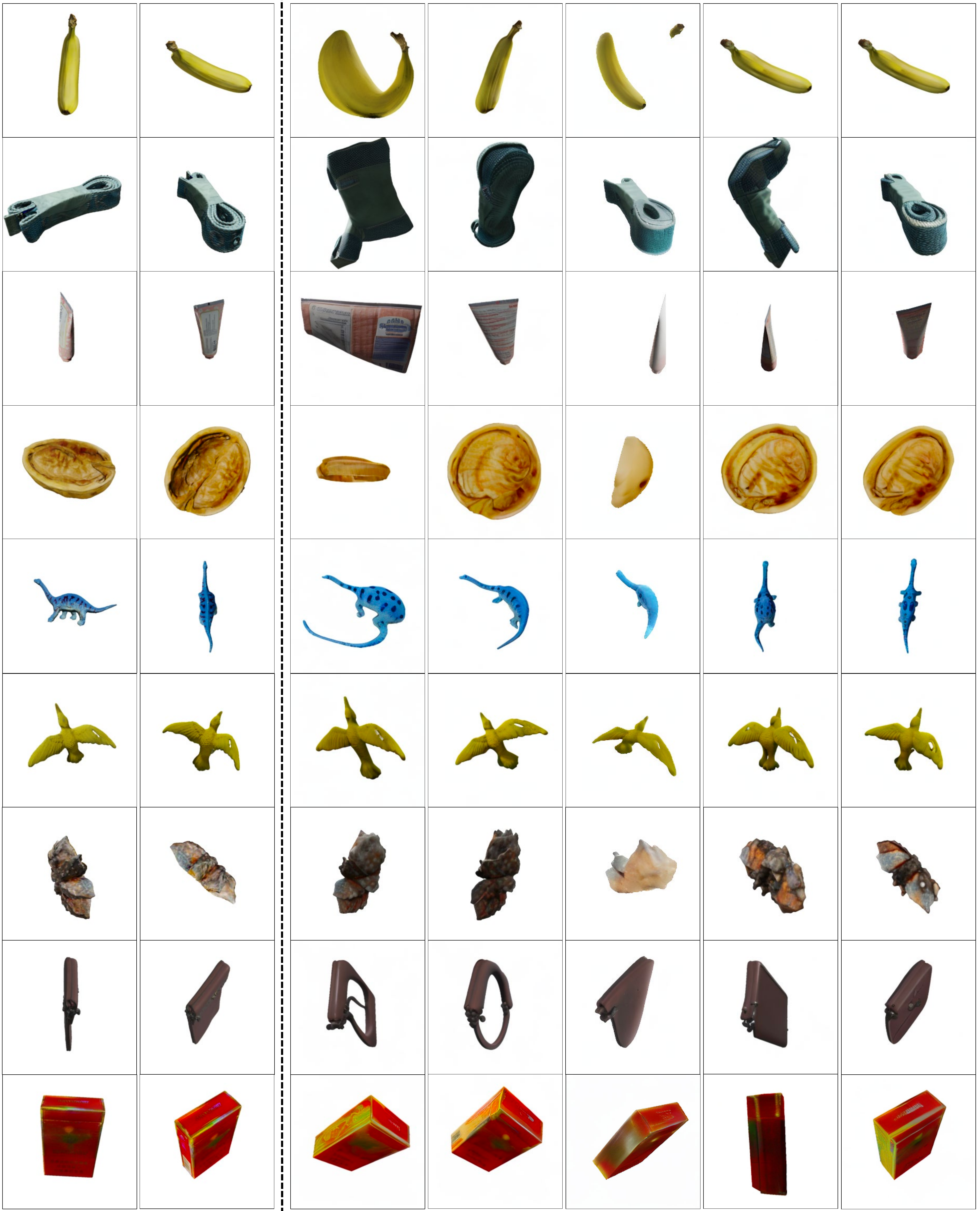}
    \begin{picture}(0,0)
    \put(-236,6){\footnotesize \textbf{(a)} Input View}
    \put(-165,6){\footnotesize \textbf{(b)} Target View}
    \put(-90,6){\footnotesize \textbf{(c)} Zero123\cite{liu2023zero}}
    \put(-28,6){\footnotesize \textbf{(d)} Zero123-XL\cite{objaverseXL}}
    \put(40,6){\footnotesize \textbf{(e)} SyncDreamer\cite{liu2023syncdreamer}}
    \put(112,6){\footnotesize \textbf{(f)} Consistent123\cite{weng2023consistent123}}
    \put(188,6){\footnotesize \textbf{(g)} Ours \method}
    \end{picture}
    \vspace{-10pt}
    \caption{\textbf{Qualitative comparisons on \texttt{OminiObject3D} dataset.}
    Given the exact target pose, the proposed \method significantly improves the pose precision compared to existing state-of-the-art methods.
    }
    \label{fig:oo3d_gso_app1}
\end{figure*}

\begin{figure*}[tb!]
    \centering
    \includegraphics[width=\linewidth]{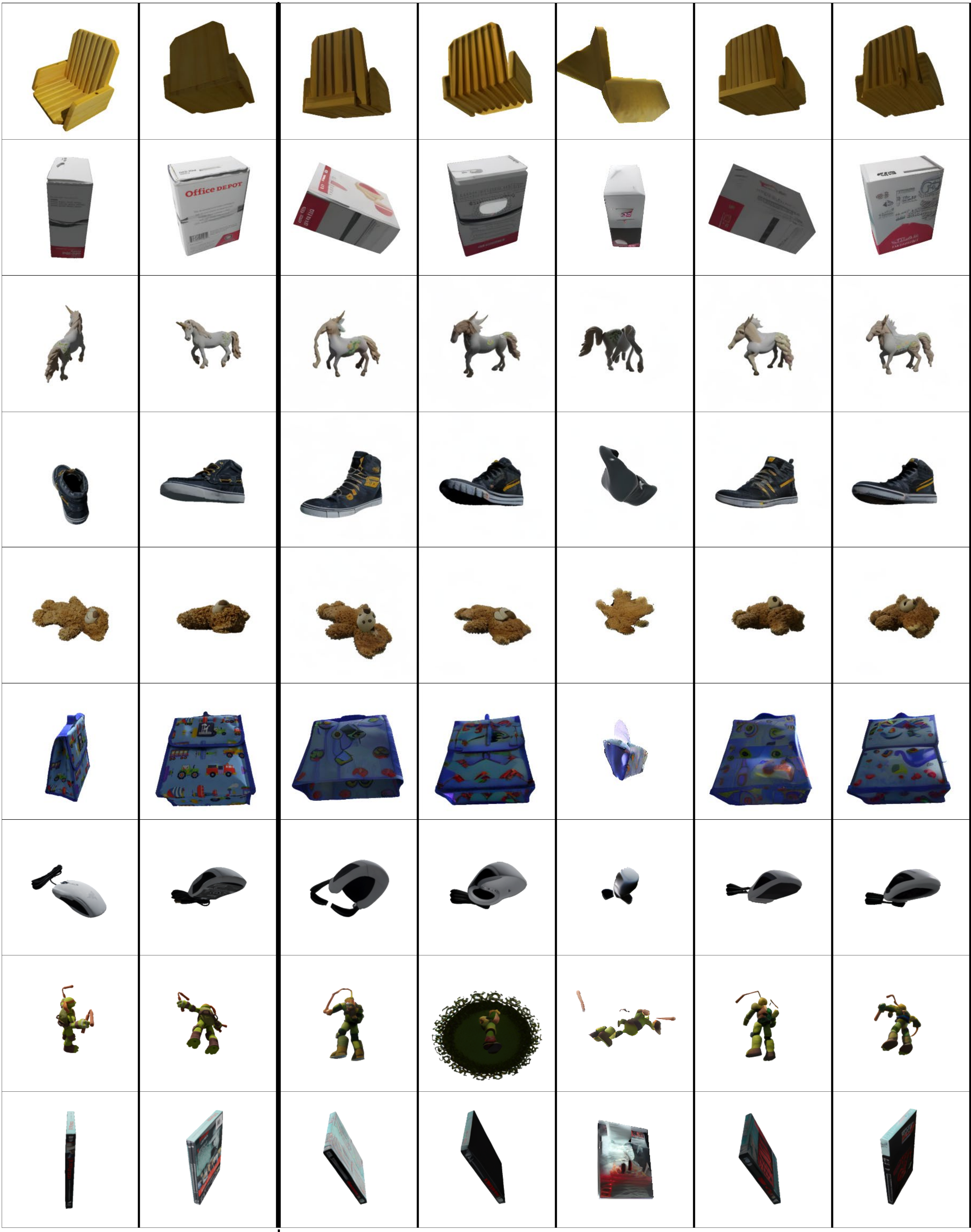}
    \begin{picture}(0,0)
    \put(-236,6){\footnotesize \textbf{(a)} Input View}
    \put(-165,6){\footnotesize \textbf{(b)} Target View}
    \put(-90,6){\footnotesize \textbf{(c)} Zero123\cite{liu2023zero}}
    \put(-28,6){\footnotesize \textbf{(d)} Zero123-XL\cite{objaverseXL}}
    \put(40,6){\footnotesize \textbf{(e)} SyncDreamer\cite{liu2023syncdreamer}}
    \put(112,6){\footnotesize \textbf{(f)} Consistent123\cite{weng2023consistent123}}
    \put(188,6){\footnotesize \textbf{(g)} Ours \method}
    \end{picture}
    \vspace{-10pt}
    \caption{\textbf{Qualitative comparisons on \texttt{GSO} dataset.}
    Given the exact target pose, the proposed \method significantly improves the pose precision compared to existing state-of-the-art methods.
    }
    \label{fig:oo3d_gso_app2}
\end{figure*}

\paragraph{More results on OmniObject3D and GSO NVS.} 
In \cref{fig:oo3d_gso_app1,fig:oo3d_gso_app2}, we show additional comparison results on \texttt{OmniObject3D}~\cite{wu2023omniobject3d} and \texttt{GSO}~\cite{downs2022google} datasets, respectively. 
This is an extension of \cref{fig:oo3d_gso} in the main paper, which demonstrates the generalizability of our \method on unseen datasets encompassing various categories.

As can be seen from these results, although the baseline Zero-1-to-3~\cite{liu2023zero} provides visually realistic appearances for all objects, the content is \emph{not} always reasonable, and the pose is inaccurate in many cases.
This indicates the global language token embedding with elevation $\theta$, azimuth $\phi$, and distance $z$ is \emph{not} so precise for the network to interpret and utilize the camera viewpoints.
While the Zero123-XL~\cite{objaverseXL} and consistent123~\cite{weng2023consistent123} enhance the quality by training on a larger dataset and employing multi-view diffusion, respectively, they do \emph{not} directly deal with the camera pose perception.
In contrast, our \method leverages the \emph{per-pixel} ray conditioning as well as the modulating, which significantly improves the pose perception accuracy.

\end{document}